\documentclass[11pt]{article}

\usepackage[final]{acl}

\usepackage{times}
\usepackage{latexsym}

\usepackage[T1]{fontenc}

\usepackage[utf8]{inputenc}

\usepackage{microtype}

\usepackage{inconsolata}

\usepackage{booktabs}   
\usepackage{makecell}   
 \usepackage{amsmath,amssymb}
 \usepackage{multirow}
\usepackage{listings}
\usepackage{tcolorbox}
\tcbuselibrary{listings}
 \newtcblisting{codeblock}{
  listing only,
  colback=black,
  coltext=green!70,
  listing options={
    basicstyle=\ttfamily\small,
    breaklines=true,       
    breakatwhitespace=true 
  },
  boxrule=0pt,
  sharp corners
}
\usepackage{amsfonts}       
\usepackage{nicefrac}       

\usepackage{xcolor}         

\usepackage{graphicx}
\usepackage{wrapfig}
\usepackage{tabularx}
\usepackage{tabularx}
\usepackage{subcaption} 
\usepackage{adjustbox}
\usepackage[ruled,vlined]{algorithm2e}
\usepackage{soul}
\sethlcolor{yellow!80}  

\title{GCoT-Decoding: Unlocking Deep Reasoning Paths for Universal Question Answering}
\author{  
Guanran Luo,  
Wentao Qiu,  
Zhongquan Jian,   
{\bf Meihong Wang, Qingqiang Wu} \\   
School of Informatics, Xiamen University \\  
\texttt{luoguanran@stu.xmu.edu.cn, wuqq@xmu.edu.cn}
}

\begin{document}
\maketitle
\begin{abstract}
Chain-of-Thought (CoT) reasoning can enhance large language models (LLMs), but it requires manually designed prompts to guide the model. Recently proposed CoT-decoding enables the model to generate CoT-style reasoning paths without prompts, but it is only applicable to problems with fixed answer sets.
To address this limitation, we propose a general decoding strategy—GCoT-decoding—that extends applicability to a broader range of question-answering tasks. GCoT-decoding employs a two-stage branching method combining Fibonacci sampling and heuristic error backtracking to generate candidate decoding paths. It then splits each path into a reasoning span and an answer span to accurately compute path confidence, and finally aggregates semantically similar paths to identify a consensus answer, replacing traditional majority voting.
We conduct extensive experiments on six datasets covering both fixed and free QA tasks. Our method not only maintains strong performance on fixed QA but also achieves significant improvements on free QA, demonstrating its generality.
\end{abstract}
\section{Introduction}
Chain-of-thought (CoT) prompting is a simple but powerful way
to elicit multi-step reasoning from large language models (LLMs),
and can substantially improve benchmark performance
\citep{kojima2022large,wei2022chain,yao2024large,yasunaga2023large,zhou2022least,lightman2023let,uesato2022solving,xie2023self,golovneva2023pathfinder}.
Most prior work, however, operates at the \emph{prompt} level:
carefully engineered instructions and exemplars are used to steer models
towards explicit CoT traces.
Such prompts inherit the designer's biases and often need to be re-tuned
across tasks and output formats
\citep{wang2022rationale,ye2022unreliability,zhou2022large}.
A complementary line of work instead modifies the \emph{decoding} process,
for example via self-consistency \citep{wang2022self},
contrastive decoding \citep{li2022contrastive},
or context-aware decoding \citep{shi2024trusting},
but these methods typically rely on extra signals
and still assume relatively rigid answer formats.

This motivates a natural question:
\emph{Can we explore and select CoT reasoning paths purely from the geometry
of the base model's decoding process, without task-specific prompts or fixed
answer formats?}
CoT-decoding \citep{wang2024chain} is an important first step.
It perturbs the first decoding step by sampling the top-$k$ alternatives,
greedily rolls out a CoT trace from each seed, extracts an answer span,
and uses the average top-1 vs.\ top-2 logit gap on that span as a
path-level confidence score.
Paths that lead to the same span are aggregated, and the answer with the
highest cumulative confidence is returned.

\begin{table}[t]
\centering
\caption{Comparison of CoT-decoding in free-form vs.\ fixed-format QA tasks.}
\label{tab:cot-free-vs-fixed}
\begingroup
\tiny              
\renewcommand{\arraystretch}{1}
\setlength{\tabcolsep}{3pt}
\begin{tabularx}{\columnwidth}{>{\raggedright\arraybackslash}p{1.5cm} X X}
\toprule
 & \textbf{Free QA} & \textbf{Fixed QA} \\
\midrule
\textbf{Example} 
& \textbf{Q}: What do Woodrow Wilson, George W.\ Bush, and James Monroe have in common? \newline
\textbf{k=1}: They all served as \textbf{presidents of the United States}. \newline
\textbf{k=2 }:\quad They were all \textbf{American leaders} involved in major wars. \newline
\textbf{k=3}: Each of them occupied the White House as \textbf{U.S.\ president}.
& \textbf{Q}: A factory makes 3 toys per hour. How many toys after 8 hours? \newline
\textbf{k=1}: $3 \times 8 = \textbf{24}$ (0.93) \newline
\textbf{k=2}: 3 times 8 is \textbf{24} (0.91) \newline
\textbf{k=3}: = \textbf{24} (0.85) \\
\midrule
\textbf{Answer Space} & $\infty$ & $\mathbb{N}$ \\
\midrule
\textbf{Exact Span Match} & $\times$ & $\checkmark$ \\
\midrule
\textbf{Majority Vote Aggregation} & $\times$ & $\checkmark$ \\
\bottomrule
\end{tabularx}
\endgroup
\end{table}

While effective on fixed-format QA, this procedure hinges on two assumptions:
(i) that a canonical answer span can be extracted reliably, and
(ii) that branching only at the first decoding step and exploring seeds in
index order is sufficient.
Table~\ref{tab:cot-free-vs-fixed} illustrates both limitations.
On a fixed-answer question (right),
all high-likelihood CoT paths end with the same numeric span ``24'',
so exact span matching and majority voting are straightforward.
On a free-form QA question (left),
the correct answer is phrased in several semantically equivalent ways,
and another path mentions a plausible but wrong alternative;
there is no unique span for aggregation by exact match.
Moreover, as our empirical analysis shows (see Appendix~\ref{app:motivation}),
early high-probability seeds often form clusters of very similar
yet incorrect continuations, while correct CoT paths are buried deeper in the ranked list.

In this paper, we revisit CoT-decoding from a decoding-time perspective
and organize the design space into three questions:
(1) \emph{Exploration}: how to spend a small path budget on diverse but
plausible reasoning directions rather than near-duplicate early seeds?
(2) \emph{Confidence}: how to score paths without relying on a single,
task-specific answer span, in both fixed-answer and free-form settings?
(3) \emph{Aggregation}: how to robustly pool free-form answers so that tiny
logit differences between nearly equivalent paths do not cause unstable
predictions?

To answer these questions, we propose
\textbf{General Chain-of-Thought Decoding (GCoT-decoding)},
a modular three-layer decoding framework (Section~\ref{sec:method}).
At the \emph{exploration layer}, GCoT combines Fibonacci-based seeding
with a single local-minimum backtracking step to allocate a small path
budget to both diverse global starts and locally “failing’’ regions of
the decoding trajectory.
At the \emph{confidence layer}, it views each path as a
``reasoning trace + answer continuation’’ and assigns a length-aware
top-2 logit gap score, with an optional LCS-based SpanAlign variant
that focuses on the aligned answer segment.
At the \emph{aggregation layer}, GCoT applies greedy semantic clustering
over answer strings and selects the representative from the cluster with
the highest accumulated confidence, aggregating paraphrase-equivalent
paths without relying on a fixed answer format.

We evaluate GCoT-decoding on six datasets spanning fixed-answer and free-form QA.
GCoT matches or slightly improves over standard multi-path decoders on
fixed-answer tasks, and yields consistent gains on free-form benchmarks
where span-based CoT-decoding struggles.
It also composes cleanly with few-shot CoT prompting and reasoning-tuned
models, providing additional improvements on top of strong baselines.
Ablation studies isolate the role of each layer and show that
Fibonacci-based multi-path exploration together with greedy semantic
clustering accounts for most of the gains, with local-minima backtracking
providing a smaller but consistent refinement. We support these design choices with an empirical analysis of
span sensitivity and exploration failure modes for CoT-decoding,
which we report in Appendix~\ref{app:motivation}.

Overall, our contributions are:
\begin{itemize} 
  \item \textbf{Proposing GCoT-Decoding}: A novel and general decoding strategy that does not rely on specific answer spans, thereby improving adaptability to diverse question-answering tasks.
  
  \item \textbf{Optimizing the branching strategy}: By introducing a two-stage branching mechanism, our method more efficiently discovers correct answers hidden in later decoding steps while correcting potentially erroneous paths.
  
  \item \textbf{Efficient path aggregation method}: We adopt a semantic similarity–based clustering strategy with a fixed threshold, and select the earliest path in each cluster as the representative. Compared to using the cluster centroid or the most similar path, this design simplifies computation while maintaining performance.

\end{itemize}

\begin{figure*}[htbp]
  \centering
  \includegraphics[width=0.86\textwidth]{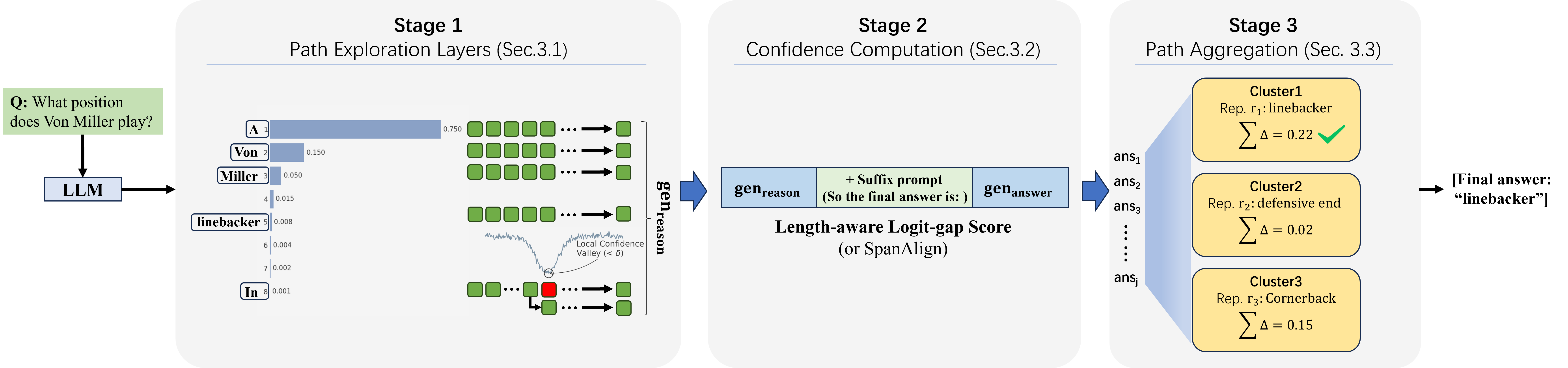}
  \caption{Overall workflow of GCoT-decoding.
  A small number of reasoning paths are first generated via Fibonacci
  seeding and confidence-based backtracking, then scored by
  length-aware logit gaps, and finally aggregated through greedy
  semantic clustering over answer strings.}
  \label{fig:2}
\end{figure*}

\section{Related Work}
\paragraph{Chain-of-Thought.}
Chain-of-Thought (CoT) prompting decomposes complex tasks into intermediate reasoning steps and has inspired a series of automated and structured extensions, including Auto-CoT, Synthetic Prompting, Contrastive Denoising CoT, Faithful CoT, and KG-CoT, which aim to improve generation quality and logical fidelity \citep{wei2022chain, kojima2022large, zhang2022automatic, shao2023synthetic, zhou2024can, lyu2023faithful, zhao2024kg}. Self-Consistency further enhances performance by aggregating diverse reasoning paths \citep{wang2022self, wang2024chain}. However, most prompting-based methods rely heavily on labeled examples, handcrafted templates, or predefined outputs, limiting scalability. In contrast, our GCoT-Decoding removes these dependencies to enable broader applicability.
\paragraph{Prompting Methods to Enhance Reasoning.}
Efforts to improve prompting strategies include paraphrasing, active example selection, analogical cues, and instruction tuning \citep{chen2024self, diao2023active, yasunaga2023large, zhang2024pattern, ho2022large}. Recent work also explores context-aware decoding and weakly-supervised aggregation to improve robustness \citep{shi2024trusting, ling2023deductive, arora2022ask}, though such methods often introduce additional annotation or computation costs. Prompt sensitivity and task specificity remain common bottlenecks.
\paragraph{Decoding Strategies to Enhance Reasoning.}
Beyond prompting, decoding-time strategies provide an alternative route for eliciting reasoning. Early contrastive decoding diversified outputs without relying on prompts \citep{li2022contrastive, yao2024large}, while self-evaluation, confidence-based scoring, and preference-guided optimization have been proposed to refine multi-step reasoning \citep{xie2023self, wang2024soft, taubenfeld2025confidence, zhang2024chain}. Tree-of-Thoughts \citep{NEURIPS2023_271db992} and CoT-decoding \citep{wang2024chain} treat reasoning as a structured exploration process, with the latter showing that top-$k$ sampling alone can reveal rich reasoning paths. Speculative decoding methods improve efficiency but are less focused on reasoning quality \citep{chen2025spin, xu2025specee}. A recent survey by \cite{welleck2024decoding} provides a comprehensive overview of decoding strategies for reasoning tasks.

\section{Method}
\label{sec:method}
Given a question $x$ and a language model
$p_\theta(\mathbf{y}\mid x)$, our goal is to uncover correct but non-prominent chain-of-thought (CoT) reasoning paths without assuming a
fixed answer format.
We organize
GCoT-decoding as a \emph{three-layer} process:
(i) a coarse path exploration layer that seeds a small number of diverse reasoning directions;
(ii) a local error-repair layer that backtracks from confidence valleys; and
(iii) an answer aggregation layer that pools evidence across paraphrase-equivalent answers.

Figure~\ref{fig:2} summarizes the workflow.
Sec.~\ref{sec:branching} describes the branching strategy for
constructing candidate paths, Sec.~\ref{sec:confidence} defines our
length-aware confidence scores and the SpanAlign variant, and
Sec.~\ref{sec:clustering} introduces greedy semantic clustering for
path aggregation.

\subsection{Two-stage branching for path exploration}
\label{sec:branching}
Our analysis in Sec.~2.2 shows that, under standard CoT-decoding, the ranked candidate list in the first few decoding steps often collapses into \emph{clusters} of very similar but incorrect reasoning trajectories. Spending a small path budget on the first few indices therefore risks exploring many near-duplicates of the same erroneous hypothesis.

GCoT-decoding instead adopts a two-stage branching scheme:
a global layer that performs \emph{one-step diversification + greedy rollout} using Fibonacci indices, and a local layer that performs \emph{backtracking from the first confidence minimum} along each path.

\paragraph{Layer 1: Fibonacci-seeded greedy rollouts.}
Let $\tilde{y}_1^{(1)}, \tilde{y}_1^{(2)}, \dots$ denote candidate tokens at the first decoding step, sorted by $p_\theta(y_1\mid x)$ in descending order. Rather than taking the first $K$ candidates, we choose indices from the Fibonacci sequence
\begin{equation}
\resizebox{\linewidth}{!}{%
  $
  S_{\mathrm{fib}}
  = \{F_1, F_2, \dots, F_K\},\qquad
  F_n = F_{n-1} + F_{n-2},\;
  F_1 = 1,\; F_2 = 2
  $
}
\end{equation}
and initialize $K$ reasoning seeds $\tilde{y}_1^{(F_1)}, \dots, \tilde{y}_1^{(F_K)}$. For each seed, the remaining tokens are generated by greedy decoding, yielding a set of candidate CoT paths $\{p_k\}_{k=1}^K$, where $p_k = (y_{k,1},\dots,y_{k,T_k})$.

Fibonacci indices implement a roughly
\emph{log-spaced coverage} of the rank axis:
for large $n$, the ratio $F_{n+1}/F_n$ approaches the golden ratio,
so seeds are increasingly spread out among lower-ranked candidates.
Under the assumption that erroneous hypotheses tend to occupy a
bounded contiguous prefix of the rank list, this log-spaced sampling
reduces the probability that all $K$ paths remain trapped in the same
error cluster.

\paragraph{Layer 2: backtracking from local confidence minima.}
Even with diverse seeds, a greedy rollout can drift toward an incorrect
answer. Empirically, such drifts are accompanied by sharp local drops in token-level confidence along the path.
We treat these as \emph{first-hitting times} of a low-confidence
region and use them as signals for local error repair.

For a greedy path $p_k = (y_{k,1}, \dots, y_{k,T_k})$, define the
token-level confidence
\begin{equation}
  s_{k,t} = p_\theta(y_{k,t} \mid x, y_{k,<t}),
  \qquad t=1,\dots,T_k.
\end{equation}
Starting from $t = 3$, we collect indices that are strict local minima
below a threshold $\delta$:
\begin{equation}
\resizebox{\linewidth}{!}{%
$
S_k = \bigl\{\,t \;\big|\; 3 \le t \le T_k,\;
s_{k,t} < s_{k,t-1},\;
(t < T_k \Rightarrow s_{k,t} < s_{k,t+1}),\;
s_{k,t} < \delta
\,\bigr\},
$
}
\end{equation}
and define the backtracking index
\begin{equation}
b_k =
\begin{cases}
\min S_k, & S_k \neq \varnothing,\\
-1, & S_k = \varnothing.
\end{cases}
\end{equation}
If $b_k \neq -1$, we step back to $y_{k,b_k-1}$ and re-branch on
$K'$ alternatives (again using Fibonacci indices over the candidate
list at that position),
\begin{equation}
\resizebox{\linewidth}{!}{%
$
\mathbf{y}_{k,<b_k}^{(m)}
= (y_{k,1}, \dots, y_{k,b_k-2},\, y_{k,b_k-1}^{(m)}),
\qquad m \in \{F_1,\dots,F_{K'}\},
$
}
\end{equation}
and complete each prefix with greedy decoding to obtain new paths
$\{p_{k,m}\}_{m=1}^{K'}$.
If $S_k = \varnothing$, we simply keep $p_k$.

This design has two effects.
First, it focuses additional computation precisely at early
confidence valleys, where the probability trajectory
$(s_{k,t})_{t}$ indicates that the current semantic direction has
fallen off the model’s high-confidence manifold.
Second, it avoids perturbing every token position.
Pseudocode for the two-stage branching scheme is provided in
Appendix~\ref{appendix:1}.

\subsection{Length-aware logit-gap confidence}
\label{sec:confidence}
After generating a set of candidate reasoning paths, we must assign a
scalar confidence score to each path for later aggregation.
Standard CoT-decoding for fixed-answer tasks often uses the average
difference between the top-1 and top-2 logits over the answer span as
a confidence proxy~\citep{wang2024chain}, but this requires a
pre-defined answer span and is brittle when answer formats vary.

GCoT-decoding replaces this with a two-step scoring scheme:
(i) a \emph{length-aware} logit-gap score based on an explicit split
between reasoning and answer segments, and
(ii) an optional \emph{SpanAlign} variant that further focuses on
tokens aligned between the reasoning and answer continuations.

\textbf{Splitting reasoning and answer segments.}
For each path $p_k$, we first let the model produce a full CoT
reasoning segment $\mathrm{gen}_{1,k}$.
We then append a short continuation prompt such as
\textit{``So the answer is:''} and decode a concise answer segment
$\mathrm{gen}_{2,k}$.
This template is used purely as a \emph{post-hoc answer extractor}
after the reasoning is complete and does not affect how the CoT itself
is generated; in Appendix~\ref{appendix:template_ablation} we show that
replacing it with semantically equivalent phrases leads to only minor variation in accuracy.

CoT paths that explore deeper reasoning directions tend to have longer
$\mathrm{gen}_{1,k}$ and more coherent answer segments.
Motivated by this, we define the base confidence of path $k$ as
\begingroup
\setlength{\abovedisplayskip}{6pt}      
\setlength{\abovedisplayshortskip}{4pt} 
\setlength{\belowdisplayskip}{6pt}      
\begin{equation}
\resizebox{0.95\linewidth}{!}{$\begin{aligned}
\Delta^{\text{GCoT-decoding}}_{k,\text{answer}}
&=
\underbrace{\frac{\log\bigl(1 + |\mathrm{gen}_{1,k}|\bigr)}
{\max_{i \in \{1,\dots,K\}} \log\bigl(1 + |\mathrm{gen}_{1,i}|\bigr)}}_{
\text{length normalization over reasoning segments}}
\\
&\quad\times\;
\underbrace{\frac{1}{|\mathrm{gen}_{2,k}|}
\sum_{x_t \in \mathrm{gen}_{2,k}}
\bigl(p(x_t^1) - p(x_t^2)\bigr)}_{
\text{average top-2 logit gap over the answer segment}}
\end{aligned}$}
\label{eq:gcot_delta}
\end{equation}
\endgroup

where $p(x_t^1)$ and $p(x_t^2)$ denote the probabilities of the top-2
tokens at decoding step $t$ in the answer segment.
The first factor encourages paths with sufficiently long reasoning,
while the second captures how confidently the model commits to the final answer.

\textbf{SpanAlign: answer-span refinement via LCS.}
In free-form settings, the same answer phrase may appear multiple times
in the reasoning and the answer continuation, and minor tokenization or
punctuation differences can lead to noisy spans.
To reduce this sensitivity, we define a SpanAlign variant based on the
longest common subsequence (LCS) between
$\mathrm{gen}_{1,k}$ and $\mathrm{gen}_{2,k}$.
Before computing the LCS, we normalize both strings by lowercasing and
stripping pure punctuation tokens, which reduces the influence of tokenization artifacts. Let $\mathrm{LCS}(\mathrm{gen}_{1,k}, \mathrm{gen}_{2,k})
= (s_{1,1},\dots,s_{1,m};\, s_{2,1},\dots,s_{2,n})$
be the aligned subsequences, with total length $L$.
We focus on the \emph{terminal} aligned spans $s_{1,m}$ and $s_{2,n}$,
which correspond to the last shared answer phrase in the reasoning and
answer segments, and define
\begin{equation}
\resizebox{0.95\linewidth}{!}{$
\begin{aligned}
\Delta^{\text{GCoT+SpanAlign}}_{k,\text{answer}}
&=
\frac{1}{L} \Biggl(
  \sum_{x_{1,t} \in s_{1,m}}
    \bigl(p(x_{1,t}^1) - p(x_{1,t}^2)\bigr)
\\
&\qquad
  +
  \sum_{x_{2,t} \in s_{2,n}}
    \bigl(p(x_{2,t}^1) - p(x_{2,t}^2)\bigr)
\Biggr)
\end{aligned}
$}
\label{eq:spanalign_delta}
\end{equation}

This concentrates the score on the shared answer span rather than on the entire continuation.

\subsection{Greedy semantic clustering for path aggregation}
\label{sec:clustering}
If we were to select the final prediction solely by
$\max_k \Delta_{k,\text{answer}}$, small perturbations in logits could
cause large jumps in the chosen path, especially when several answers
are close in confidence.
Aggregating across multiple paths can mitigate this sensitivity, but standard majority voting is not applicable on open-ended tasks where answers are free-form text and exact string matches are rare.
We therefore aggregate at the level of \emph{semantic answer clusters}.

Let $\{p_i\}_{i=1}^K$ denote the candidate paths produced by the
branching stages, with final answers
$g_i = \mathrm{gen}_{2,i}$ and confidence scores
$c_i = \Delta_{i,\text{answer}}$.
We maintain a set of semantic groups $\{G_j\}_{j=1}^N$ with
representative answers $\{r_j\}_{j=1}^N$, initially empty.
For each answer $g_i$ in index order, we compute cosine similarities
\begin{equation}
  s_{i,j} = \cos\bigl(\phi(g_i),\, \phi(r_j)\bigr),
  \qquad j = 1,\dots,N,
\end{equation}
where $\phi(\cdot)$ is a sentence embedding function.
We then assign $g_i$ according to the greedy rule
\begin{equation}
\resizebox{\linewidth}{!}{$
j^* =
\begin{cases}
\min\{\,j \in \{1,\dots,N\}\mid s_{i,j} \ge \tau\},
& \text{if } \max_{1 \le j \le N} s_{i,j} \ge \tau,\\[4pt]
N+1, & \text{otherwise},
\end{cases}
$}
\end{equation}

which always chooses the first existing cluster above a similarity
threshold $\tau$, or creates a new cluster if none qualify.
We update
\begin{equation}
\resizebox{\linewidth}{!}{$
G_{j^*} \leftarrow G_{j^*} \cup \{g_i\},\qquad
r_{j^*} =
\begin{cases}
r_{j^*}, & j^* \le N,\\
g_i,     & j^* = N+1,
\end{cases}
\quad
N \leftarrow \max(N, j^*).
$}
\end{equation}

After all $K$ answers are processed, we compute the cumulative
confidence of each group
\begin{equation}
  C_j = \sum_{g_i \in G_j} c_i,\qquad j = 1,\dots,N,
\end{equation}
and select the representative $r_{j_{\max}}$ with
$j_{\max} = \arg\max_j C_j$ as the final output.

We further ablate the sentence embedding model and find that performance varies only slightly (Appendix~\ref{appendix:embed_ablation}), suggesting that the clustering module is relatively insensitive to the particular off-the-shelf encoder. Pseudocode for the aggregation procedure is also given in
Appendix~\ref{appendix:1}.

\section{Results and Analysis}

\begin{table*}[ht]
  \centering
    \fontsize{25}{20}\selectfont
  \renewcommand{\arraystretch}{1.4}
  \setlength{\tabcolsep}{4pt}         
  \resizebox{\textwidth}{!}{%
    \begin{tabular}{%
      l 
      >{\centering\arraybackslash}m{3em}  
      c c c  c c c  c c c
    }
      \toprule
      & \multirow{2}{*}{\shortstack{Spec\\Ans}}
        & \multicolumn{3}{c}{GSM8K}
        & \multicolumn{3}{c}{MultiArith}
        & \multicolumn{3}{c}{Sports understanding} \\
      \cmidrule(lr){3-5} \cmidrule(lr){6-8} \cmidrule(lr){9-11}
      & 
        & Mistral‑7B & Gemma‑7B & Llama‑3.1‑8B
        & Mistral‑7B & Gemma‑7B & Llama‑3.1‑8B
        & Mistral‑7B & Gemma‑7B & Llama‑3.1‑8B \\
      \midrule
      Greedy                     & ×
                                 & 10.5 & 11.6 & 17.9
                                 & 16.0 & 18.7 & 38.8
                                 & 49.6 & 61.2 & 51.6 \\
      Temperature sampling       & ×
                                 &  8.4 &  7.9 & 13.1
                                 & 15.2 & 18.8 & 36.2
                                 & 48.9 & 60.1 & 52.4\textsuperscript{$\dagger$} \\
      Top‑k sampling             & ×
                                 &  5.1 &  6.2 & 14.2
                                 & 13.3 & 17.3 & 37.0
                                 & 50.3 & 58.0 & 51.9 \\
                                 \midrule
        Beam search        & ×
                                 & 6.7 & 10.2 & 17.1
                                 & 15.5 & 17.9 & 38.1
                                 & 48.2 & 59.9 & 50.7 \\
      CoT‑decoding               & \(\checkmark\)
                                 & 21.9\textsuperscript{$\spadesuit$}
                                 & 25.4\textsuperscript{$\spadesuit$}
                                 & 36.3\textsuperscript{$\dagger$}
                                 & 40.6\textsuperscript{$\spadesuit$}
                                 & 43.8\textsuperscript{$\spadesuit$}
                                 & 72.3\textsuperscript{$\dagger$}
                                 & 50.6
                                 & 68.4\textsuperscript{$\spadesuit$}
                                 & 51.0 \\
      Self-consistency  & \(\checkmark\)
                                 & 16.3 & 17.2 & 28.5
                                 & 21.7 & 22.9 & 46.9
                                 & 52.9\textsuperscript{$\spadesuit$} & 63.9 & 54.6\textsuperscript{$\dagger$} \\
    GCoT-decoding + SpanAlign   & ×
                                 & 10.7 & 15.4 & 34.0
                                 & 16.8 & 19.7 & 69.3
                                 & 48.0 & 67.2\textsuperscript{$\dagger$} & 52.0 \\
      \textbf{GCoT‑decoding}     & ×
                                 & 18.0\textsuperscript{$\dagger$}
                                 & 21.8\textsuperscript{$\dagger$}
                                 & 41.7\textsuperscript{$\spadesuit$}
                                 & 31.3\textsuperscript{$\dagger$}
                                 & 22.8\textsuperscript{$\dagger$}
                                 & 74.3\textsuperscript{$\spadesuit$}
                                 & 52.0\textsuperscript{$\dagger$}
                                 & 65.2
                                 & 58.0\textsuperscript{$\spadesuit$} \\
      \bottomrule
    \end{tabular}%
  }
  \caption[Accuracy comparison of decoding strategies on fixed QA tasks]
  {Accuracy comparison of decoding strategies on fixed QA tasks; the top‑ranked is marked with $\spadesuit$ and the second‑ranked is marked with $\dagger$. Spec Ans indicates whether the decoding strategy relies on specific answer spans. The top section lists single-path decoding strategies; the bottom section shows multi-path decoding strategies.}
  \label{tab:fixed_qa}
\end{table*}
\begin{table*}[ht]
  \centering
    \fontsize{35}{30}\selectfont
  \renewcommand{\arraystretch}{1.5}%
  \setlength{\tabcolsep}{4pt}%
  \resizebox{\textwidth}{!}{%
    \begin{tabular}{l c c c c c c c c c c c c c c c c c c}
      \toprule
      & \multicolumn{6}{c}{SQuAD v1.1 (contextual)}
      & \multicolumn{6}{c}{BARQA (contextual)}
      & \multicolumn{6}{c}{Auto categorization (context‑free)} \\
      \cmidrule(lr){2-7} \cmidrule(lr){8-13} \cmidrule(lr){14-19}
      & \multicolumn{2}{c}{Gemma‑7B}
      & \multicolumn{2}{c}{Llama‑3.1‑8B}
      & \multicolumn{2}{c}{Qwen2.5‑14B}
      & \multicolumn{2}{c}{Gemma‑7B}
      & \multicolumn{2}{c}{Llama‑3.1‑8B}
      & \multicolumn{2}{c}{Qwen2.5‑14B}
      & \multicolumn{2}{c}{Gemma‑7B}
      & \multicolumn{2}{c}{Llama‑3.1‑8B}
      & \multicolumn{2}{c}{Qwen2.5‑14B} \\
      \cmidrule(lr){2-3} \cmidrule(lr){4-5} \cmidrule(lr){6-7}
      \cmidrule(lr){8-9} \cmidrule(lr){10-11} \cmidrule(lr){12-13}
      \cmidrule(lr){14-15} \cmidrule(lr){16-17} \cmidrule(lr){18-19}
      & BLEU & MATCH & BLEU & MATCH & BLEU & MATCH
      & BLEU & MATCH & BLEU & MATCH & BLEU & MATCH
      & BLEU & MATCH & BLEU & MATCH & BLEU & MATCH \\
      \midrule
      Greedy
        &  3.3   & 42.8   &  8.3   & 60.6   & 21.4   & 67.2
        &  4.7   & 36.6\textsuperscript{$\dagger$}   & 10.8   & 39.7   & 10.7\textsuperscript{$\dagger$}   & 44.4\textsuperscript{$\spadesuit$}
        &  5.8   & 16.8   &  5.1\textsuperscript{$\dagger$}   & 16.0\textsuperscript{$\dagger$}   &  8.5   & 29.0 \\
      Temperature sampling
        &  3.1   & 40.1   &  7.5   & 57.2   & 17.1   & 64.1
        &  4.5   & 32.1   &  7.3   & 37.4   &  7.7   & 42.5
        &  6.0   & 13.6   &  4.9   & 13.3   &  6.6   & 27.9 \\
      Top‑k sampling
        &  2.8   & 35.2   &  5.4   & 51.0   & 13.1   & 55.1
        &  2.9   & 33.3   &  6.8   & 37.2   &  6.4   & 40.0
        &  4.3   & 13.7   &  4.5   & 11.2   &  5.6   & 26.0 \\
        \midrule
    Beam Search
        &  3.2 & 41.9 &  7.9 & 59.3 & 20.0 & 66.0
        &  4.2 & 35.4 & 10.0 & 38.5 & 10.1 & 42.1
        &  5.3 & 15.0 &  4.7 & 15.4 &  8.1 & 28.4 \\
    CoT‑decoding + Prompt-based
        &  0.2 & 25.7 & 1.3 & 40.9 & 5.8 & 50.3
        &  0.7 & 21.5 & 2.4 & 25.1 & 1.4 & 32.0
        &  1.2 & 20.1 & 2.0 &  15.7 &  8.0 & 29.0 \\ 
    Self-consistency + Prompt-based
        &  4.2\textsuperscript{$\dagger$} & 36.7 &  3.2 & 43.2 & 12.1 & 58.0
        &  2.2 & 26.1 & 3.6 & 30.4 & 1.5 & 33.5
        &  7.4 & 20.3 &  3.1 & 14.1 &  5.3 & 29.8 \\
      GCoT-decoding + SpanAlign
        &  3.9 & 48.9\textsuperscript{$\dagger$}
        &  9.2\textsuperscript{$\dagger$} & 62.0\textsuperscript{$\dagger$}
        & 21.5\textsuperscript{$\dagger$} & 69.6\textsuperscript{$\dagger$}
        &  5.8\textsuperscript{$\dagger$} & 36.5
        & 10.9\textsuperscript{$\dagger$} & 41.5\textsuperscript{$\dagger$}
        & 10.9\textsuperscript{$\spadesuit$} & 43.3\textsuperscript{$\dagger$}
        &  8.8\textsuperscript{$\dagger$} & 23.3\textsuperscript{$\dagger$}
        &  4.5   & 14.7   &  8.8\textsuperscript{$\dagger$}   & 30.2\textsuperscript{$\dagger$} \\
      \textbf{GCoT‑decoding}
        &  4.9\textsuperscript{$\spadesuit$} & 54.6\textsuperscript{$\spadesuit$}
        & 10.0\textsuperscript{$\spadesuit$} & 67.2\textsuperscript{$\spadesuit$}
        & 23.2\textsuperscript{$\spadesuit$} & 71.4\textsuperscript{$\spadesuit$}
        & 10.9\textsuperscript{$\spadesuit$} & 37.7\textsuperscript{$\spadesuit$}
        & 12.3\textsuperscript{$\spadesuit$} & 44.1\textsuperscript{$\spadesuit$}
        & 10.2 & 38.9
        &  8.9\textsuperscript{$\spadesuit$} & 24.6\textsuperscript{$\spadesuit$}
        &  6.8\textsuperscript{$\spadesuit$} & 20.0\textsuperscript{$\spadesuit$}
        & 10.6\textsuperscript{$\spadesuit$} & 30.5\textsuperscript{$\spadesuit$} \\
      \bottomrule
    \end{tabular}%
  }
  \caption[Performance of different models on free QA tasks]
  {Performance of different models on free QA tasks; the top‑ranked is marked with $\spadesuit$ and the second‑ranked is marked with $\dagger$. The top section lists single-path decoding strategies; the bottom section shows multi-path decoding strategies.}
  \label{tab:free_qa}
\end{table*}

\subsection{Experimental setup}
\paragraph{Datasets.}
We evaluate models on two categories of QA tasks: (1) \emph{Fixed~QA}, where the answer set or format is constrained, including \textbf{GSM8K} and \textbf{MultiArith} \citep{cobbe2021training, roy-roth-2015-solving} for multi-step arithmetic reasoning, and \textbf{Sports understanding} \citep{suzgun2022challenging} from Big-Bench-Hard for binary reasoning over sports-related sentences; and (2) \emph{Free~QA}, which involves open-ended or paragraph-level outputs, such as \textbf{SQuAD v1.1} \citep{rajpurkar2016squad} for extractive reading comprehension, \textbf{BARQA} \citep{srivastava2022beyond} for context-dependent anaphora resolution, and \textbf{Auto Categorization} \citep{srivastava2022beyond} for identifying semantic categories among object sets.
\paragraph{Baseline Methods and Evaluation Metrics.}
We primarily compare decoding-based methods, including \textbf{single-path sampling} strategies such as greedy decoding, temperature sampling ($t = 0.7$), and top-$k$ sampling ($k = 10$); as well as \textbf{multi-path sampling} methods like beam search ($b = 10$), self-consistency ($k = 10$) \citep{wang2022self} and CoT-decoding \citep{wang2024chain}.

We do not include prompt-based methods as baselines, as they are orthogonal to GCoT-decoding and can be freely combined (see Appendix~\ref{Compatibility} for discussion). For \textbf{fixed QA}, we use \textit{accuracy}, computed by comparing the extracted answer token against the ground truth—note this extraction is used only for evaluation, not confidence computation. For \textbf{free QA}, we evaluate with \textit{BLEU} \citep{papineni2002bleu} and \textit{MATCH}, which checks whether the ground-truth span appears in the response. For GCoT-decoding variants, \textit{BLEU} is calculated only on the final answer $\mathrm{gen}_2$.

\paragraph{Model and Parameter Settings.}
In the main experiments, we evaluate four models: Mistral-7B \citep{jiang2024identifying}, Gemma-7B \citep{team2024gemma}, Llama3.1-8B \citep{grattafiori2024llama}, and Qwen2.5-14B \citep{yang2024qwen2}.  For the model-scale ablation, we use the Qwen2.5 series at 3B, 7B, 14B, and 32B scales. We use all‑MiniLM‑L6‑v2 \citep{reimers2019sentence} as the embedding model. We set the first-stage branching number $k=10$ and second-stage branching number $k'=2$, branch only when confidence falls below a threshold $\delta$ of $0.2$.  During semantic aggregation of paths, we use a similarity threshold $\tau$ of $0.8$. To ensure the stability and reliability of our findings, all results reported for the main experiments are calculated as the average of three independent runs.

\subsection{Main results}

\textbf{Fixed QA.} As shown in Table~\ref{tab:fixed_qa}, GCoT-decoding outperforms all single-path decoding strategies (greedy and sampling methods) and most multi-path decoding strategies (beam search and self-consistency) across all models and datasets. Although CoT-decoding achieves the highest accuracy on math reasoning tasks, its performance heavily relies on specific answer spans. This dependency explains its advantage in fixed QA tasks but also becomes a major bottleneck when extending to free QA tasks. In contrast, GCoT-decoding offers a more stable alternative that does not rely on answer spans, achieving competitive performance on fixed QA while delivering significant gains on free QA.

\textbf{Free QA.}
As shown in Table~\ref{tab:free_qa}, GCoT-decoding achieves the highest BLEU and MATCH scores in nearly all settings, significantly outperforming other methods in both generation quality and answer alignment. Even compared to variants such as CoT-decoding + Prompt-based and Self-consistency + Prompt-based, GCoT-decoding remains the top performer. In contrast, GCoT-decoding + SpanAlign suffers from performance drops due to frequent misalignment with incorrect spans. Overall, GCoT-decoding demonstrates stronger robustness and generality when tackling complex, free-form reasoning tasks.

\subsection{Compatibility of GCoT-decoding with Prompting Methods}
\label{Compatibility}
\begin{figure}[t]
  \centering
  \includegraphics[width=0.5\textwidth]{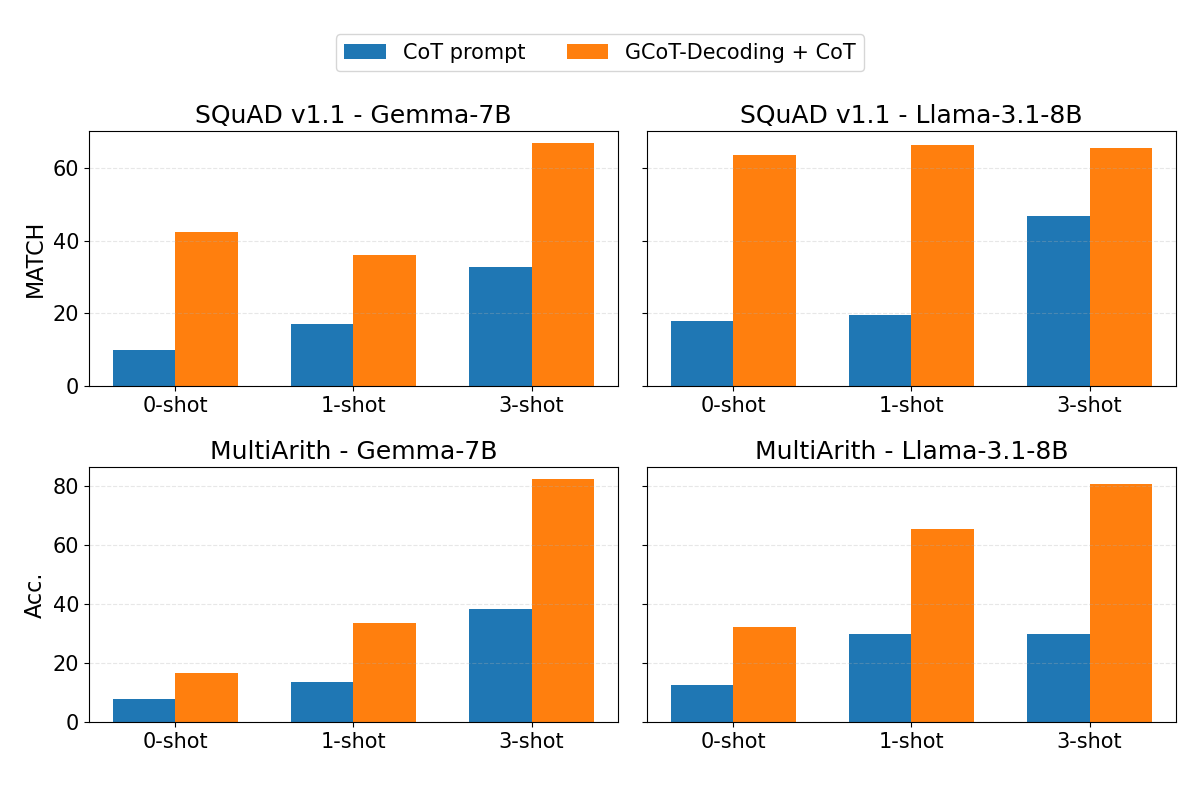}
  \caption{The results of combining GCoT-decoding with CoT prompting.}
  \label{fig:6}
\end{figure}
Although GCoT-decoding is a prompt-free method, this does not preclude its combination with prompt-based approaches; in fact, they are highly compatible. Experiments on MultiArith and SQuAD v1.1 using Gemma-7B and Llama-3.1-8B show (Figure~\ref{fig:6}) that merging GCoT-decoding with CoT prompting yields steady performance improvements across all few-shot settings in both fixed and free QA, with absolute gains of 10\%–50\%. This demonstrates that GCoT-decoding and CoT prompting synergize effectively, significantly enhancing LLM reasoning quality in few-shot scenarios. We provide the few-shot examples used in Appendix~\ref{appendix:3}.

\subsection{Ablation study}

\begin{table}[t]
  \centering
  \resizebox{\linewidth}{!}{%
    \begin{tabular}{lcccc}
      \toprule
      Variant
      & \makecell[c]{GSM8K\\(Gemma-7B)}
      & \makecell[c]{GSM8K\\(Mistral-7B)}
      & \makecell[c]{SQuAD v1.1\\(Gemma-7B)}
      & \makecell[c]{SQuAD v1.1\\(Llama-3.1-8B)} \\
      \midrule
      Fibonacci + greedy (ours)
      & 21.8 & 18.0 & 54.6 & 67.2 \\
      top-$k$ sampling ($k{=}10$)
      & 7.9  & 6.2  & 42.1 & 50.4 \\
      top-$p$ sampling ($p{=}0.9$)
      & 8.6  & 7.0  & 43.5 & 51.3 \\
      temperature sampling ($T{=}0.7$)
      & 9.4  & 7.8  & 45.0 & 52.6 \\
      \bottomrule
    \end{tabular}%
  }
  \caption{Ablation of path-generation strategies under a fixed budget of $K{=}10$ paths. All variants share the same backtracking and aggregation modules.}
  \label{tab:path_gen_ablation}

\end{table}

We ablate GCoT-decoding along its three main stages: (i) the path generation strategy, (ii) the backtracking rule, and (iii) the path aggregation module. Appendix~\ref{appendix:2} reports additional ablations, including alternative confidence computation schemes and multi-factor variants where several modules are simplified simultaneously.

\textbf{Effect of path generation strategy.} Our goal differs from generic diversity generation: instead of injecting randomness at every step, we only diversify the first token to open a few alternative reasoning directions and then greedily roll out each path. Fibonacci indices further spread this first-step sampling budget along the ranked candidates in a roughly log-spaced manner, avoiding redundant exploration of tightly clustered early hypotheses. Under a fixed budget of $K{=}10$ paths, Table~\ref{tab:path_gen_ablation} compares this Fibonacci-based scheme to standard step-wise stochastic sampling while keeping backtracking and aggregation fixed, and shows that replacing our “one-step diversification + greedy rollout’’ with top-$k$/top-$p$/temperature sampling drives GSM8K accuracy down to about 8–10\% and reduces SQuAD v1.1 MATCH by 10–20 points.

\label{sec:backtracking_ablation}

\textbf{Reliability of local-minima backtracking.} We assess reliability by measuring trigger frequency and success rate on SQuAD v1.1 (Table~\ref{tab:backtracking_ablation}). Local-minima backtracking is triggered on only about 28\% of questions, yet fixes an otherwise wrong greedy answer in 36.5\% of those cases, raising MATCH from 52.7 to 54.6. Random and late backtracking are always triggered but slightly underperform the no-backtracking baseline and have much lower conditional success rates (around 18–21\%), indicating that naive perturbations are not helpful. We further study the effect of allowing more backtracking points per path in Appendix~\ref{appendix:6}.
\begin{table}[t]
  \centering
  \resizebox{\linewidth}{!}{%
    \begin{tabular}{lcccc}
      \toprule
      Variant
      & \makecell[c]{Backtracking\\trigger rate (\%)} 
      & \makecell[c]{Success rate\\given trigger (\%)}
          & \makecell[c]{MATCH}
      & \makecell[c]{BLEU} \\
      \midrule
      No-backtracking
      & --   & --   & 52.7 & 8.7 \\
      Random backtracking
      & 100.0 & 18.1 & 52.0 & 8.6 \\
      Late backtracking
      & 100.0 & 20.4 & 51.8 & 8.5 \\
      Local-minima backtracking (ours)
      & 28.0 & 36.5 & 54.6 & 9.1 \\
      \bottomrule
    \end{tabular}%
  }
  \caption{Backtracking variants on SQuAD v1.1 dev (Gemma-7B); ``Success rate given trigger'' is the fraction of triggered cases corrected by backtracking.}
  \label{tab:backtracking_ablation}

\end{table}

\begin{table}[t]

  \centering
  \small
  \resizebox{\linewidth}{!}{%
    \begin{tabular}{lccc}
      \toprule
      Aggregation variant
      & \makecell[c]{Extra time\\per question (sec.)}
      & \makecell[c]{GSM8K\\Acc.\ (Gemma-7B)}
      & \makecell[c]{SQuAD\\MATCH (Gemma-7B)} \\
      \midrule
      MaxPath (no aggregation)
      & 0.0 & 15.3 & 41.9 \\
      Greedy clustering (ours)
      & 0.2 & 21.8 & 54.6 \\
      LLM-based aggregation
      & 8.3 & 22.1 & 55.8 \\
      \bottomrule
    \end{tabular}%
  }
  \caption{MaxPath vs.\ greedy semantic clustering and an LLM-based aggregation module (Gemma-7B). Extra time is measured relative to greedy decoding.}
  \label{tab:aggregation_ablation}

\end{table}

\begin{figure*}[htbp]
  \centering
  \begin{minipage}[t]{0.28\textwidth}
    \centering
    \includegraphics[height=1.8cm]{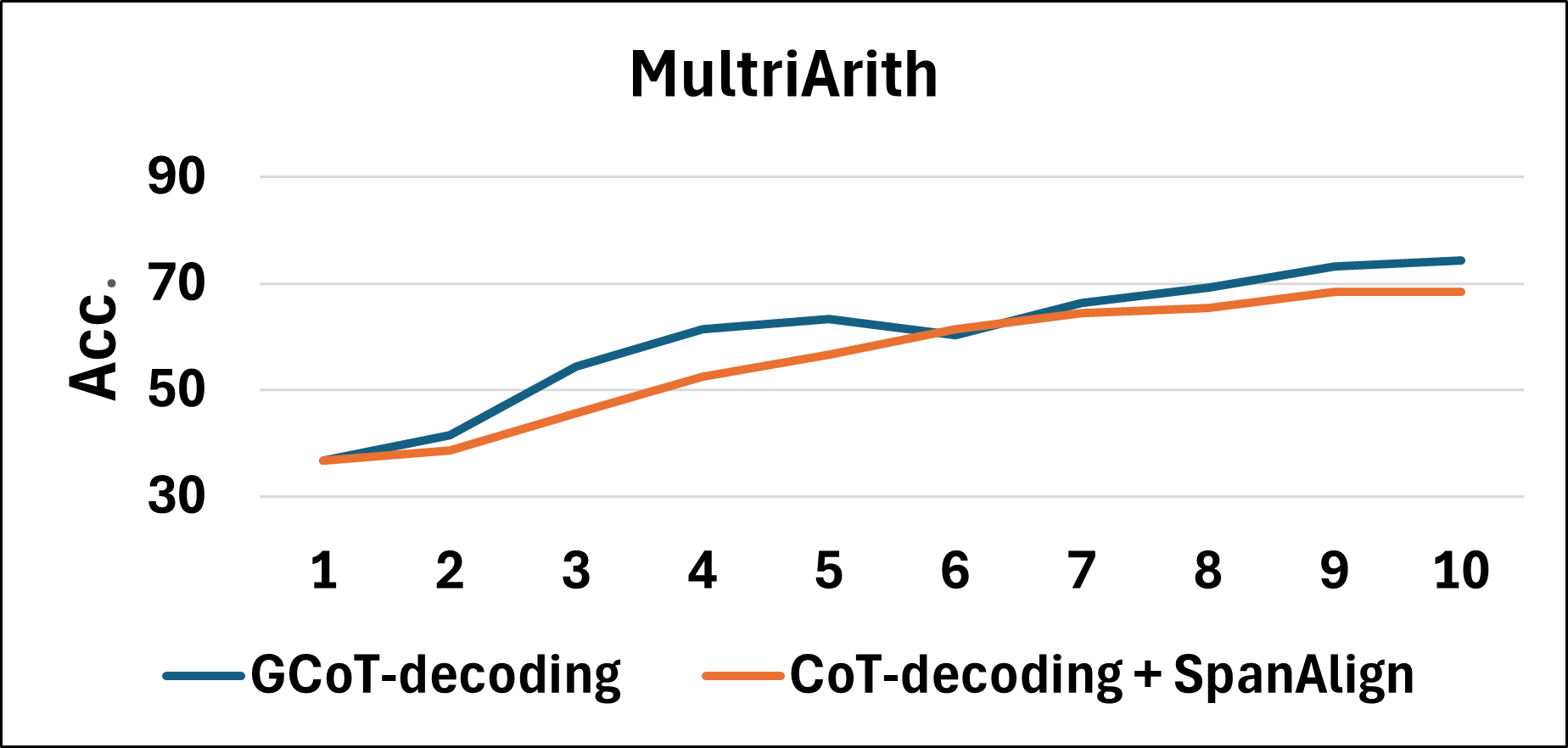}\\[-0.1em]
    \includegraphics[height=1.8cm]{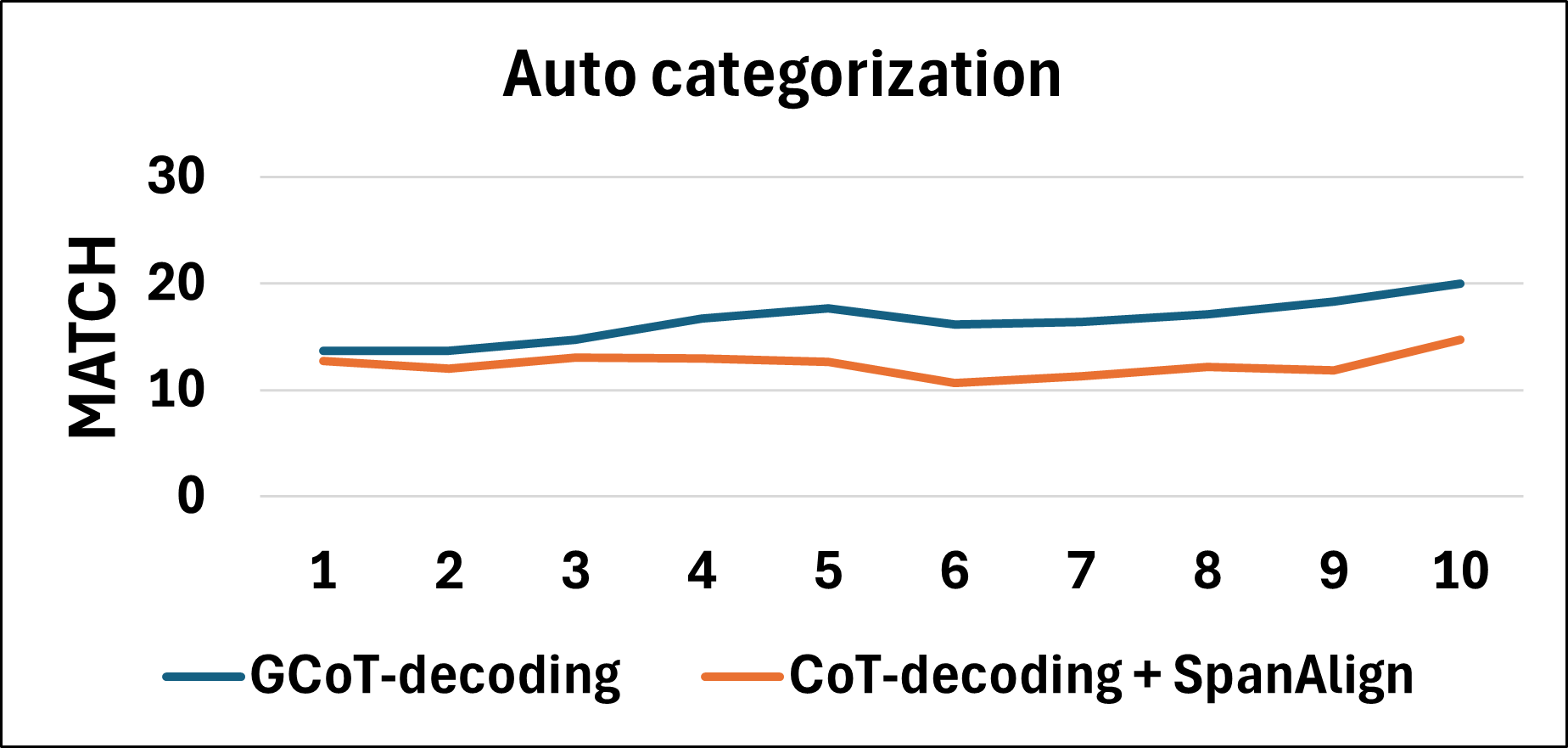}
    \caption*{(a) Effect of model size.}
  \end{minipage}%
  \hspace{1.05em}
  \begin{minipage}[t]{0.28\textwidth}
    \centering
    \includegraphics[height=1.8cm]{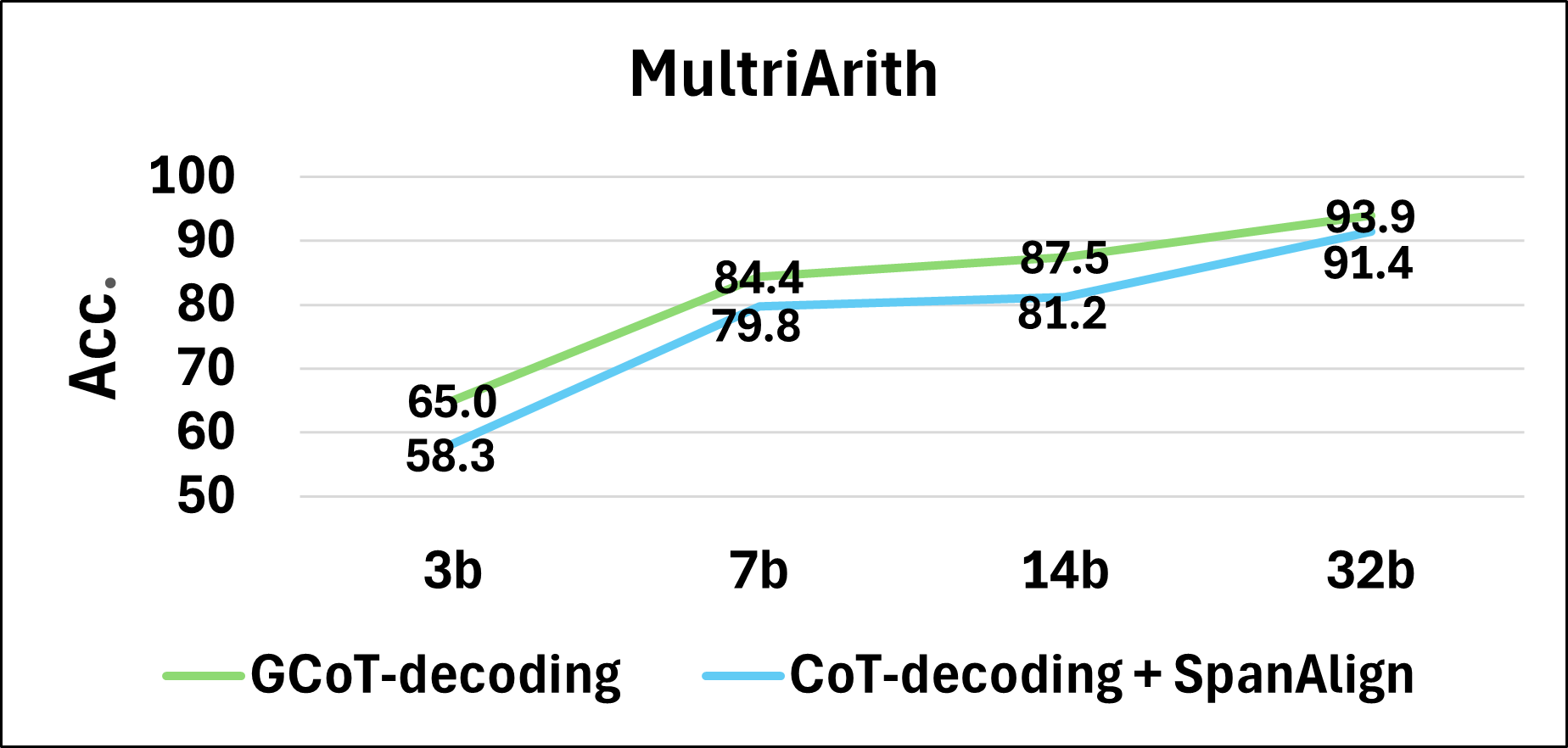}\\[-0.1em]
    \includegraphics[height=1.8cm]{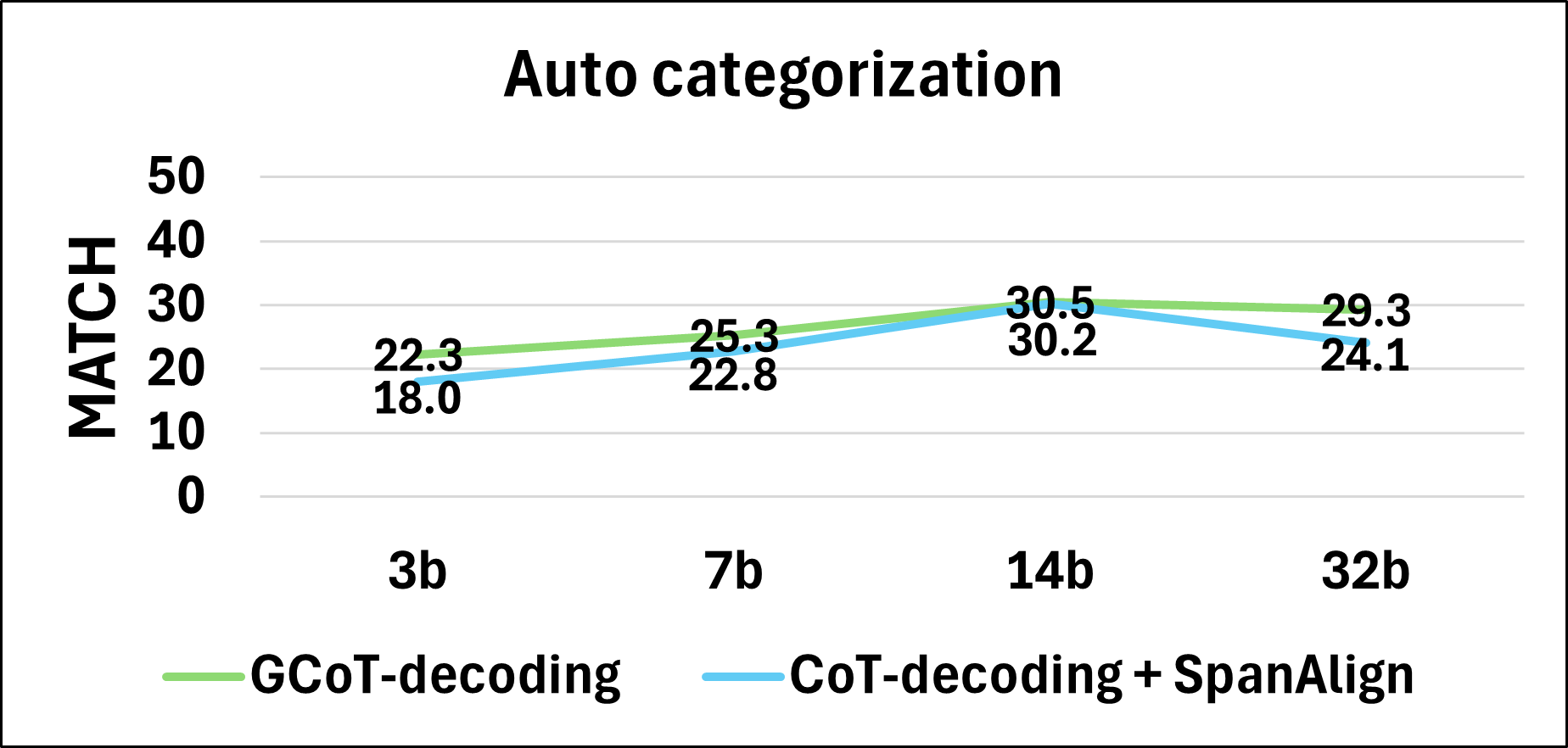}
    \caption*{(b) Impact of decoding path count \(k\).}
  \end{minipage}
  \hspace{0.8em}
  \begin{minipage}[t]{0.28\textwidth}
    \centering
    \includegraphics[height=1.8cm]{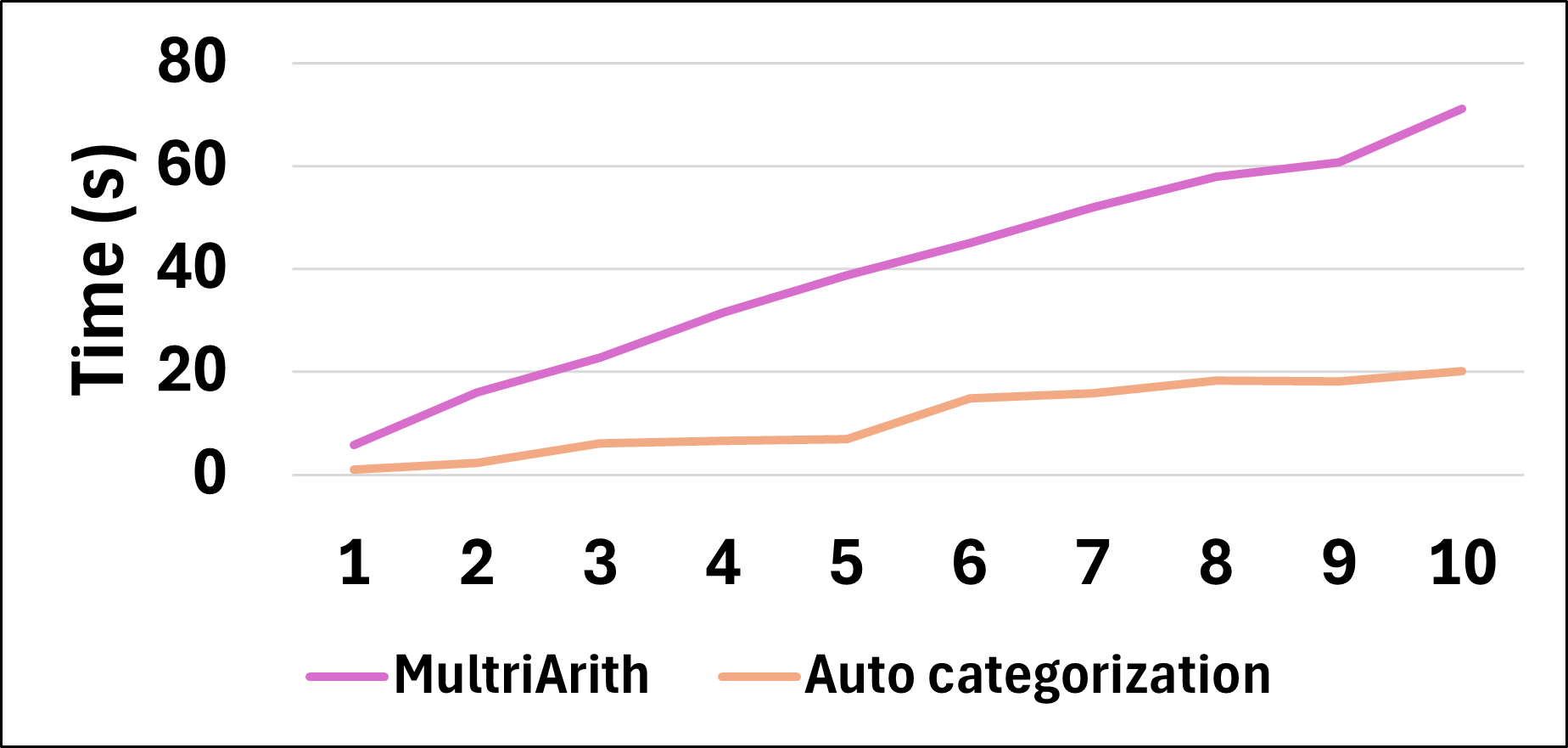}\\[-0.1em]
    \includegraphics[height=1.8cm]{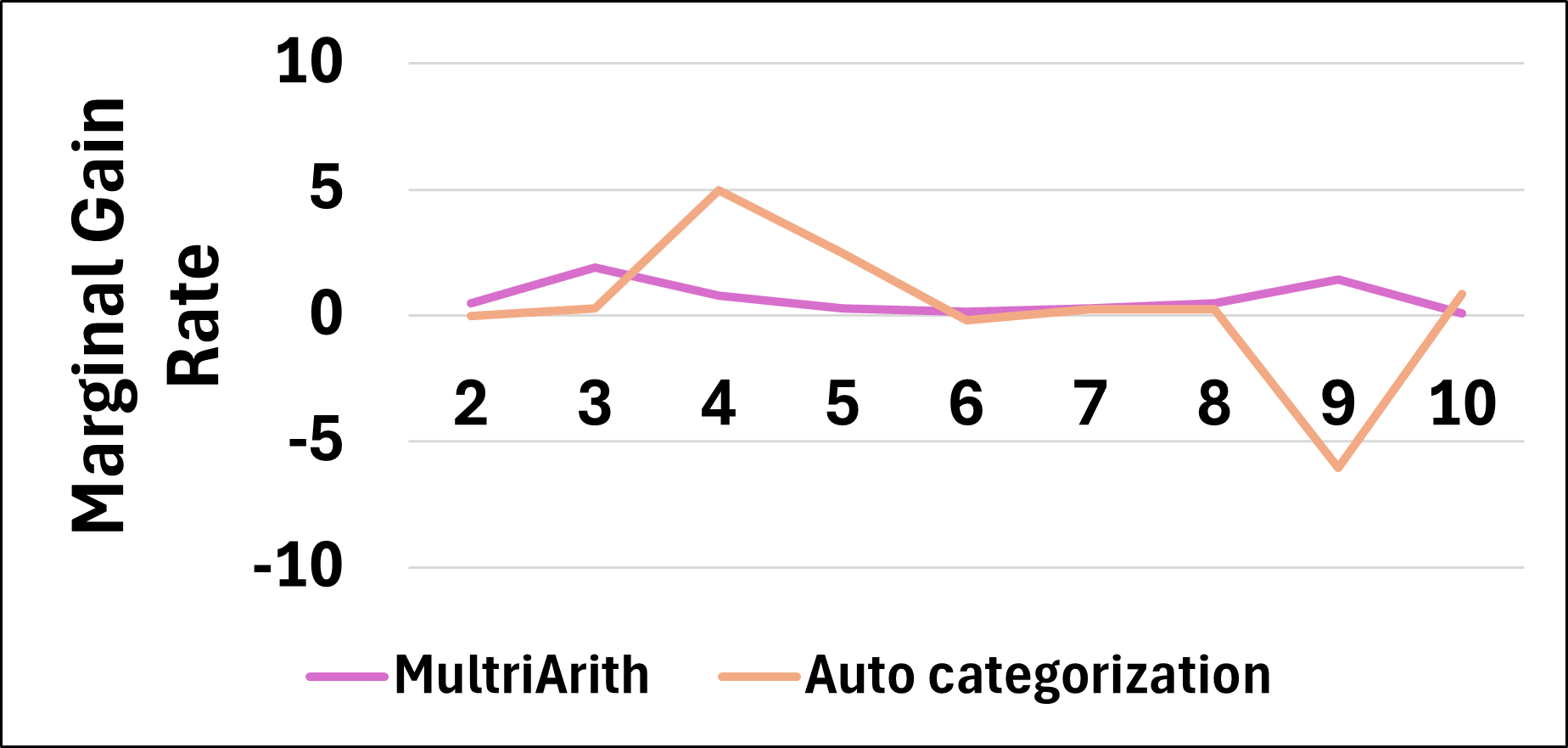}
    \caption*{(c) Time cost and marginal gain rate by decoding path count \(k\).}
  \end{minipage}
  \caption{The impact of model size and the number of decoding paths \(k\).}
  \label{fig:wrap-merged}
\end{figure*}

\begin{table*}[htbp]
\centering
\tiny
\setlength{\tabcolsep}{4pt}
\renewcommand{\arraystretch}{1.25}

\caption{Decoding outputs with confidence gaps $\Delta_{k,\text{answer}}$ for two classification examples.}
\label{tab:merged-path-examples}
\begin{tabular}{c p{0.42\textwidth} c p{0.42\textwidth} c}
\toprule

& \textbf{Question:} AUSTRO-ITALIAN WAR, JACOBITE REBELLION, and FRANCO-SPANISH WAR are instances of
&  
& \textbf{Question:} Profitable home Chelisheva, The House with Lions, and House under the steeple can be classified as
&  \\
\midrule

& \textbf{Ground truth:} historical wars
&  
& \textbf{Ground truth:} tourist attractions / architecture in Russia
&  \\
\midrule
k=1
& \textit{European diplomatic initiatives.} So the answer is: \textbf{European diplomatic initiatives} (\(\Delta\)=0.22) & × 
& \textit{These are notable tourist attractions located across Russia.} So the answer is: \textbf{tourist attractions} (\(\Delta\)=0.81) & \checkmark \\
k=2
& \textit{diplomatic initiatives.} So the answer is: \textbf{diplomatic initiatives.} (\(\Delta\)=0.18) & × 
& \textit{architectural heritage in Russia.} So the answer is: \textbf{architecture in Russia} (\(\Delta\)=0.68) & \checkmark \\
k=3
& \textit{These events can be categorized under diplomatic initiatives.} So the answer is: \textbf{diplomatic initiatives} (\(\Delta\)=0.09) & × 
& \textit{tourist attractions in Russia. Explanation: each of these locations is a notable architectural site known for its historical significance within Russian cities.} So the answer is: \textbf{tourist attractions} (\(\Delta\)=0.93) & \checkmark \\
k=5
& \textit{They are relevant to international treaty formation.} So the answer is: \textbf{international treaty formation} (\(\Delta\)=0.14) & × 
& \textit{They refer to government-owned residential complexes.} So the answer is: \textbf{government-owned residential complexes} (\(\Delta\)=0.24) & × \\
k=8
& \textit{Historical wars, because each conflict exemplifies armed struggles ...} So the answer is: \textbf{historical wars} (\(\Delta\)=0.81) & \checkmark 
& \textit{metaphors from Soviet-era literature about class struggle.} So the answer is: \textbf{Soviet-era literature} (\(\Delta\)=0.11) & × \\
\bottomrule
\end{tabular}
\end{table*}

\textbf{Greedy semantic clustering vs.\ LLM-based aggregation.} We first compare GCoT-decoding with a MaxPath baseline that simply selects the single highest-confidence path: as shown in Table~\ref{tab:aggregation_ablation}, greedy semantic clustering improves GSM8K accuracy from 15.3 to 21.8 and SQuAD MATCH from 41.9 to 54.6, with only 0.2 seconds of extra time per question. An LLM-based aggregator yields slightly higher scores than greedy clustering but incurs about 8.3 seconds of additional latency and is sensitive to the aggregation prompt. Our greedy clustering therefore offers most of the aggregation benefit over MaxPath at a fraction of the compute cost, matching our goal of a lightweight, robust aggregation module.

\subsection{Quantitative and qualitative analysis}

\textbf{Quantitative analysis.}  
As shown in Figure~\ref{fig:wrap-merged}(a), performance improves with scale, especially from 3B to 7B, with smaller gains beyond. \emph{GCoT‐decoding} consistently outperforms \emph{+SpanAlign} across scales and shows greater robustness to domain shifts. Figure~\ref{fig:wrap-merged}(b) shows that increasing the number of decoding paths \(k\) initially improves performance but saturates after \(k>5\). \emph{GCoT‐decoding} maintains stronger and more stable gains than \emph{+SpanAlign} across all \(k\) settings. As shown in Figure~\ref{fig:wrap-merged}(c), time cost grows roughly linearly with \(k\), while both tasks exhibit diminishing marginal gains. Taken together, the optimal “elbow” lies in the range \(k=3\sim5\), where the marginal gain rate peaks and time remains moderate.

\textbf{How Fibonacci sampling works.}
Table~\ref{tab:merged-path-examples} shows two case studies of Fibonacci sampling.
In the war classification example, paths $k=1$–$3$ all converge on related but wrong labels such as “diplomatic initiatives,” and the correct label “historical wars” only emerges at $k=8$, with a clear reasoning chain—illustrating how early high-probability seeds can cluster around the same mistake.
Here, Fibonacci sampling skips over these local error clusters and reaches the correct path with fewer probes.
In the architecture example, where the top-ranked path is already correct, early paths ($k=1$–$3$) also yield correct labels (with $k=3$ providing a particularly explicit explanation); even though some later correct paths are skipped, the correct label remains dominant in the aggregated confidence.

\begin{figure}[t]
  \centering
  \includegraphics[width=1\linewidth]{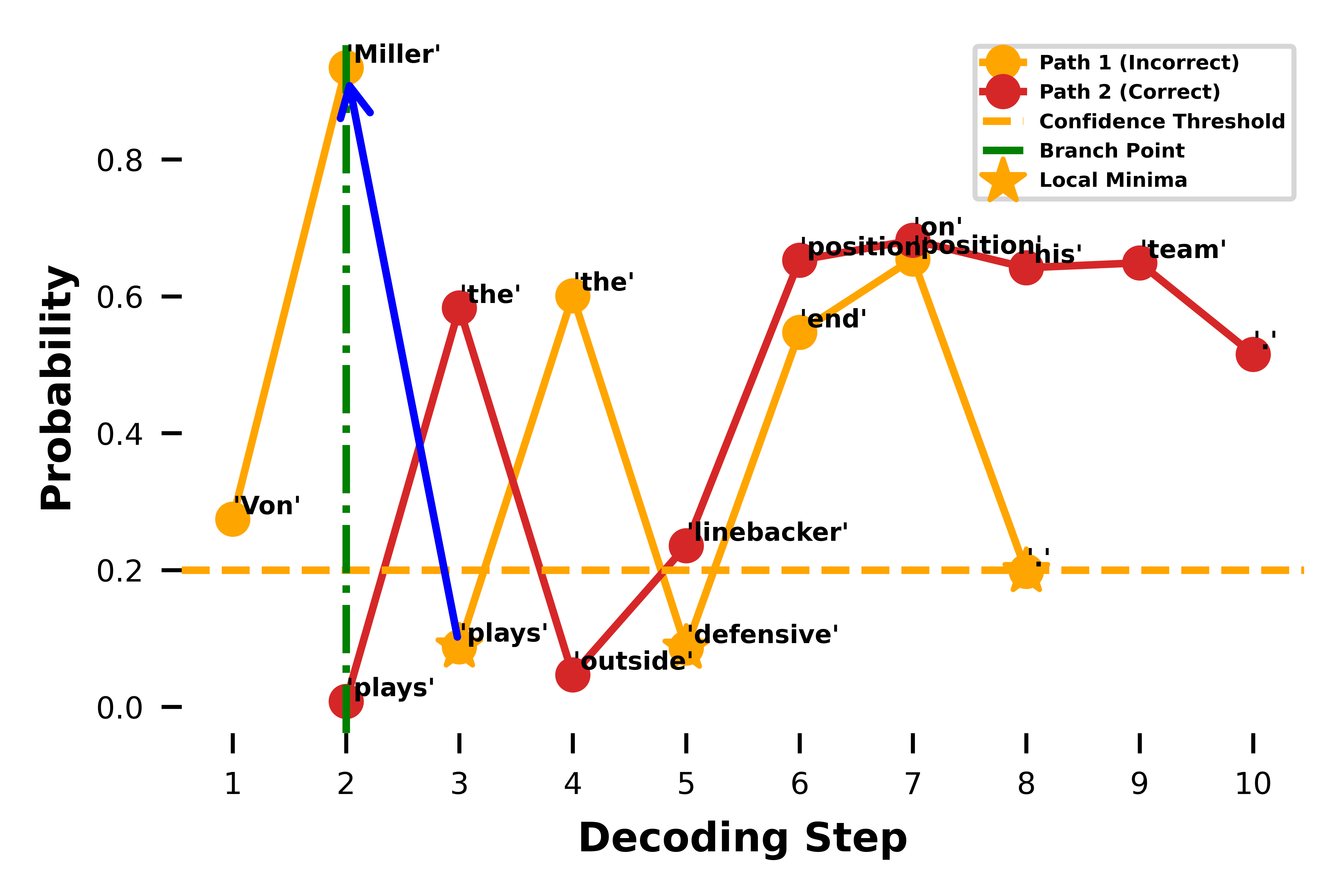}
\caption{Illustration of early path backtracking.}
  \label{fig:error-entrenchment}
\end{figure}

\textbf{How early path backtracking works.} 
Figure~\ref{fig:error-entrenchment} illustrates how early backtracking prevents errors from becoming entrenched.
In Path~1 (orange), the model drifts toward the incorrect span \textit{``defensive end''}, with several local minima falling below the confidence threshold (yellow stars).
Our rule treats these minima as warning signals and branches before the first one (at step~2), creating Path~2 (red), which instead converges to the correct answer \textit{``linebacker''}.
Once the erroneous span has been fully generated and reinforced by high-confidence tokens, later branching rarely fixes it, underscoring the importance of backtracking early.

\section{Conclusion}
We propose GCoT-Decoding, a general decoding strategy that extends earlier chain-of-thought based methods to broader QA tasks. By refining the branching mechanism for generating candidate paths, our approach further boosts performance. Experiments show that GCoT-Decoding consistently improves the reasoning ability of language models of various sizes and offers greater robustness to task drift across diverse benchmarks.

\section*{Limitations}
Despite these benefits, GCoT-Decoding introduces additional computational overhead due to exploring and maintaining multiple reasoning paths. Our current evaluation is also limited to a set of QA and reasoning benchmarks, and does not fully cover tasks where reasoning is more implicit (e.g., summarization-style generation). Going forward, we are exploring optimizations such as early path pruning and more adaptive branching to reduce computational cost, as well as extending evaluation to a wider range of tasks that require step-by-step reasoning (e.g., structured text generation, logical inference, and multi-hop reasoning). For summarization-like tasks, we plan to investigate hybrid approaches that selectively apply GCoT-Decoding only to reasoning-intensive components, aiming to balance efficiency with broader applicability.


\bibliography{custom}

\appendix

\section{Additional Analysis of CoT-decoding}
\label{app:motivation}

\subsection{Sensitivity of CoT-decoding to answer-span extraction}
\label{app:span-sensitivity}

\begin{figure*}[t]
\centering
\includegraphics[width=0.85\textwidth]{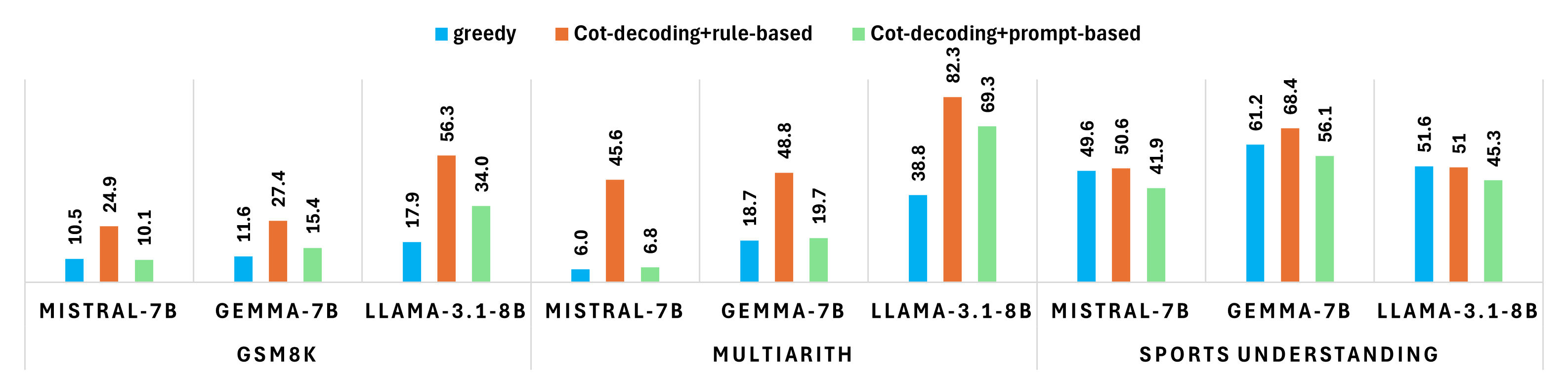}
\caption{Impact of different answer extraction strategies on CoT-decoding performance.}
\label{fig:motivation}
\end{figure*}

A natural way to extend CoT-decoding~\citep{wang2024chain} beyond fixed-answer
math questions is to ask the model to restate the final answer, e.g., by
appending a marker such as \textit{``So the answer is:''} and computing
confidence on the continuation.
However, when the same answer phrase appears multiple times in the trace,
or when the wording of the marker varies slightly, different choices of
``answer span'' can lead to noticeably different confidence scores.

We compare two extraction methods on GSM8K, MultiArith, and the BBH
\emph{Sports Understanding} benchmark.
The \textbf{rule-based} extractor follows the official evaluation
protocols on these fixed-answer tasks:
for GSM8K and MultiArith it takes the last number in the response,
and for \emph{Sports Understanding} it takes the final binary token
(yes/no).
The \textbf{prompt-based} extractor instead extends the model output
with \textit{``So the answer is:''} and uses the continuation as the
answer span.
As shown in Figure~\ref{fig:motivation}, simply switching from the
rule-based span to the prompt-based span can substantially degrade
CoT-decoding: on GSM8K and MultiArith it often collapses toward the
greedy baseline, and on \emph{Sports Understanding} it yields 5--12
point drops.

These results highlight two issues.
First, CoT-decoding is structurally tied to a single, task-specific
answer span, which limits its applicability to free-form QA where no
canonical span exists.
Second, even when such a span is available, small changes to how it is
identified can have a surprisingly large impact.
This motivates the confidence layer of GCoT-decoding
(Section~\ref{sec:confidence}), where we treat each path as a
``reasoning trace + answer continuation'' rather than relying on a
hand-picked span.

\subsection{Greedy exploration can miss deeper correct paths}
\label{app:greedy-exploration}

\begin{table}[t]
  \centering
  \caption{Distribution of correct and incorrect paths and their corresponding confidences for the top 100 GSM8K questions in the case of first-index error.}
  \label{tab:motivation2}
  \tiny
  \begin{tabular}{ccccc}
    \toprule
    \textbf{Index} & \textbf{Correct} & \textbf{Incorrect} & \textbf{C. Conf.} & \textbf{I. Conf.} \\
    \midrule
    0   & --  & --  & --   & --   \\
    1   & 8   & 92  & 0.73 & 0.09 \\
    2   & 2   & 98  & 0.68 & 0.13 \\
    3   & 13  & 87  & 0.70 & 0.10 \\
    4   & 23  & 77  & 0.74 & 0.14 \\
    $\cdots$ & $\cdots$ & $\cdots$ & $\cdots$ & $\cdots$ \\
    9   & 44  & 56  & 0.62 & 0.20 \\
    \bottomrule
  \end{tabular}
\end{table}

We also examine how CoT-decoding explores candidate paths.
In the original formulation, paths are generated by perturbing only the
first decoding step and then greedily rolling out each seed.
Candidate paths are ordered by likelihood, and naive multi-path
strategies simply take the first $K$ of them.
This can be a poor search strategy when early high-probability seeds
form clusters of near-duplicate yet incorrect continuations.

Following \citet{wang2024chain}, we analyze the top 100 GSM8K questions
under a controlled setting where the first-ranked path is incorrect.
For each index $i \in \{1,\dots,9\}$, we measure
(i) how often the path at index $i$ is correct vs.\ incorrect, and
(ii) the average confidence of correct and incorrect paths at that index.
Table~\ref{tab:motivation2} summarizes the distribution of correctness
and confidence across indices.
When the first index is wrong, the next few indices are also dominated
by incorrect paths with similar surface forms and intermediate reasoning
steps.
Correct paths only become common at larger indices, so sweeping many
adjacent early indices yields diminishing returns while consuming a
large decoding budget.

This observation motivates the exploration layer of GCoT-decoding
(Section~\ref{sec:branching}):
rather than exhaustively sweeping consecutive indices, we use Fibonacci
indices to spread a fixed seed budget roughly log-uniformly over the
ranked candidates, and add a local-minimum backtracking mechanism that
spawns a few additional branches when the token-level confidence along a
path exhibits a sharp local drop.
Together, these mechanisms allow GCoT to reach deeper correct paths more
efficiently than naive index-ordered exploration.

\section{Algorithm Details}
\label{appendix:1}
We provide the pseudocode of path sampling and backtracking in Algorithm 1, and the pseudocode of the decoding path aggregation algorithm based on semantic clustering in Algorithm 2.
\begin{algorithm}
\small
\SetAlgoLined
\KwIn{Model \texttt{model}, tokenizer \texttt{tokenizer}, query \texttt{query}, first branching size $k$, second branching size $k'$, confidence threshold $\delta$}
\KwOut{List of final decoding paths}

Initialize empty result list $\mathcal{R}$\;

// First branching
Compute logits from initial \texttt{query} using \texttt{model}\;

Select tokens at indices determined by Fibonacci sequence: $\{F_1, F_2, \dots, F_k\}$\;

\ForEach{token index $i \in \{F_1, F_2, \dots, F_k\}$}{
    Form initial decoding prefix by appending token $t_i$ to \texttt{query}\;
    Greedily decode from this prefix to obtain complete path $\mathbf{y}=(y_1,y_2,\dots,y_T)$ and token confidences $\{s_1,s_2,\dots,s_T\}$\;
    Append path and confidences to temporary list $\mathcal{L}$\;
}

// Secondary branching via backtracking
\ForEach{decoded path $\mathbf{y}$ and confidences $\{s_t\}_{t=1}^T$ in $\mathcal{L}$}{
    Identify local minima set $S=\{t\mid 3\le t\le T, s_t<s_{t-1},(t<T \Rightarrow s_t<s_{t+1}), s_t<\delta\}$\;

    Determine branching point $b$:
    \[
    b = \begin{cases}
    \min S, & S\ne\varnothing \\
    -1, & S=\varnothing
    \end{cases}
    \]

    \If{$b \ne -1$}{
        Truncate path to form prefix $\mathbf{y}_{<b}=(y_1,\dots,y_{b-2})$\;
        Compute logits for next token after prefix $\mathbf{y}_{<b}$\;
        Select alternative tokens at Fibonacci indices $\{F_1,\dots,F_{k'}\}$\;

        \ForEach{alternative token index $j \in \{F_1,\dots,F_{k'}\}$}{
            Append token $y_{b-1}^{(j)}$ to prefix $\mathbf{y}_{<b}$\;
            Greedily decode from new prefix to complete new path $\mathbf{y}^{(j)}$\;
            Add new path $\mathbf{y}^{(j)}$ to result list $\mathcal{R}$\;
        }
    }
    \Else{
        Add original path $\mathbf{y}$ directly to result list $\mathcal{R}$\;
    }
}

\Return result list $\mathcal{R}$
\caption{General decoding path generation with Fibonacci sampling and backtracking}
\label{alg:fib-backtracking}
\end{algorithm}
\begin{algorithm}
\small
\SetAlgoLined
\KwIn{Decoding paths \(\{p_i\}_{i=1}^K\), confidences \(\{c_i\}_{i=1}^K\), embedding function \(\phi(\cdot)\), similarity threshold \(\tau\)}
\KwOut{Final aggregated answer}

Initialize semantic groups: \(G_j \leftarrow \emptyset\), representatives \(r_j \leftarrow \emptyset\), group count \(N \leftarrow 0\)\;

\ForEach{path output \(g_i = \mathrm{gen}_2(p_i)\)}{
    Compute embedding \(\phi(g_i)\)\;
    \If{N = 0}{
        Create new group \(G_1 = \{g_i\}\), set representative \(r_1 = g_i\), set \(N=1\)\;
        \textbf{continue}\;
    }

    Compute similarities \(s_{i,j} = \cos(\phi(g_i), \phi(r_j))\) for all existing groups \(j = 1, \dots, N\)\;

    Find the minimal index \(j^*\) satisfying \(s_{i,j^*} \geq \tau\); if none exist, set \(j^* = N + 1\)\;

    \eIf{\(j^* \leq N\)}{
        Add \(g_i\) to existing group \(G_{j^*}\)\;
    }{
        Create new group \(G_{N+1} = \{g_i\}\), set representative \(r_{N+1} = g_i\), increment \(N\)\;
    }
}

Compute cumulative confidence \(C_j = \sum_{g_i \in G_j} c_i\) for each group \(j\)\;

Select group with maximum cumulative confidence \(j_{\max} = \arg\max_j C_j\)\;

Return group representative \(r_{j_{\max}}\) as the final output.
\caption{General decoding path aggregation via semantic clustering}
\label{alg:path-aggregation}
\end{algorithm}
We provide the pseudocode of path sampling and backtracking in Algorithm 1, and the pseudocode of the decoding path aggregation algorithm based on semantic clustering in Algorithm 2.

\section{Extended Ablation and Hyperparameter Sensitivity}
\label{appendix:2}

\subsection{Extended Ablation Study of GCoT-Decoding}
\label{app:extended_ablation}

\begin{table*}[htbp]
  \centering
  \resizebox{\textwidth}{!}{%
  \begin{tabular}{l l l c c c c}
    \toprule
    Confidence Computation & Path Generation & Path Aggregation 
    & \makecell[c]{GSM8K\\(Acc., Gemma-7B)} 
    & \makecell[c]{GSM8K\\(Acc., Mistral-7B)} 
    & \makecell[c]{SQuAD v1.1\\(MATCH, Gemma-7B)} 
    & \makecell[c]{SQuAD v1.1\\(MATCH, Llama-3.1-8B)} \\
    \midrule
    --        & --         & --        & \textbf{21.8} & \textbf{18.0} & \textbf{54.6} & \textbf{67.2} \\
    entropy   & --         & --        & 18.3 (–3.5)   & 14.1 (–3.9)   & 51.4 (–3.2)   & 63.0 (–4.2)   \\
    logits    & --         & --        & 19.0 (–2.8)   & 15.7 (–2.3)   & 54.5 (–0.1)   & 66.9 (–0.3)   \\
    --        & Seq        & --        & 17.5 (–4.3)   & 13.1 (–4.9)   & 48.9 (–5.7)   & 63.6 (–3.6)   \\
    --        & OneBranch  & --        & 20.8 (–1.0)   & 16.5 (–1.5)   & 52.7 (–1.9)   & 67.0 (–0.2)   \\
    --        & --         & MaxPath   & 15.3 (–6.5)   & 12.8 (–5.2)   & 41.9 (–12.7)  & 50.8 (–16.4)  \\
    \midrule
    entropy   & Seq        & --        & 13.9 (–7.9)   & 11.2 (–6.8)   & 42.1 (–12.5)  & 49.4 (–17.8)  \\
    logits    & --         & MaxPath   & 12.5 (–9.3)   & 10.7 (–7.3)   & 39.7 (–14.9)  & 45.6 (–21.6)  \\
    --        & Seq        & MaxPath   & 12.2 (–9.6)   & 9.9 (–8.1)    & 36.3 (–18.3)  & 44.7 (–22.5)  \\
    entropy   & --         & MaxPath   & 12.8 (–9.0)   & 10.3 (–7.7)   & 40.8 (–13.8)  & 46.5 (–20.7)  \\
    --        & OneBranch  & MaxPath   & 11.5 (–10.3)  & 8.8 (–9.2)    & 33.6 (–21.0)  & 41.9 (–25.3)  \\
    entropy   & Seq        & MaxPath   & 8.4 (–13.4)   & 6.7 (–11.3)   & 25.8 (–28.8)  & 33.0 (–34.2)  \\
    logits    & Seq        & MaxPath   & 8.9 (–12.9)   & 7.4 (–10.6)   & 27.5 (–27.1)  & 34.6 (–32.6)  \\
    entropy   & OneBranch  & MaxPath   & 7.7 (–14.1)   & 5.4 (–12.6)   & 19.5 (–35.1)  & 24.8 (–42.4)  \\
    \bottomrule
  \end{tabular}%
  }
  \caption{Ablation study of GCoT-Decoding. Top rows show single-factor ablations; bottom rows show selected multi-factor variants. Numbers in parentheses denote drops from the full model.}
  \label{tab:ablation_refined}
\end{table*}

As shown in Table~\ref{tab:ablation_refined}, computing path confidence using the softmax probability gap between the top-2 tokens consistently outperforms raw logits and entropy across tasks. While raw logits are sensitive to distributional shifts—especially in open-ended QA—entropy tends to misrepresent confidence due to token fragmentation in LLMs. For path generation, replacing Fibonacci sampling with sequential decoding reduces the likelihood of reaching correct answers, and removing the second-stage backtracking prevents correction of low-confidence tokens, allowing flawed reasoning paths to persist. Moreover, selecting only the highest-confidence path (MaxPath) significantly undermines decoding stability; on SQuAD, this leads to performance drops of up to 16.4\%. These results underscore the importance of multi-path aggregation in mitigating single-path errors and capturing diverse yet valid reasoning chains, which are essential for robust GCoT-Decoding. Overall, when multiple components are simultaneously simplified, the performance of GCoT-Decoding deteriorates rapidly, underscoring the importance of all three modules working in concert.

\subsection{Sensitivity to Hyperparameters}

We also provide sensitivity experiments on the similarity threshold $\tau$ and the confidence threshold $\delta$, summarized in Table~\ref{tab:tau-delta-results}.

\begin{table*}[h]
\centering
\caption{Performance under different thresholds $\tau$ and $\delta$ on GSM8K, MultiArith, and Sports Understanding tasks.}
\label{tab:tau-delta-results}
\resizebox{\textwidth}{!}{%
\begin{tabular}{cc|ccc|ccc|ccc}
\toprule
$\tau$ & $\delta$ 
& \multicolumn{3}{c|}{GSM8K} 
& \multicolumn{3}{c|}{MultiArith} 
& \multicolumn{3}{c}{Sports Underst.} \\
\cmidrule(lr){3-5} \cmidrule(lr){6-8} \cmidrule(lr){9-11}
& & Mistral-7b & Gemma-7b & Llama-3.1-8b 
  & Mistral-7b & Gemma-7b & Llama-3.1-8b 
  & Mistral-7b & Gemma-7b & Llama-3.1-8b \\
\midrule
0.8 & 0.2 & 18.0 & 21.8 & 41.7 & 31.3 & 23.2 & 74.3 & 52.0 & 65.2 & 58.0 \\
0.7 & 0.2 & 16.9 & 20.5 & 40.8 & 30.1 & 21.9 & 72.6 & 49.8 & 63.7 & 55.3 \\
0.9 & 0.2 & 17.3 & 21.0 & 40.9 & 30.5 & 22.7 & 73.2 & 51.5 & 64.2 & 57.0 \\
0.8 & 0.1 & 17.2 & 21.1 & 41.1 & 30.7 & 22.4 & 73.4 & 51.7 & 64.5 & 57.3 \\
0.8 & 0.3 & 17.4 & 21.4 & 41.2 & 30.6 & 22.6 & 73.7 & 51.6 & 64.7 & 57.5 \\
\bottomrule
\end{tabular}%
}
\end{table*}

\section{Prompt Demonstration Examples}
\label{appendix:3}
\begin{figure}[h]
  \centering
  \resizebox{\linewidth}{!}{%
    \includegraphics{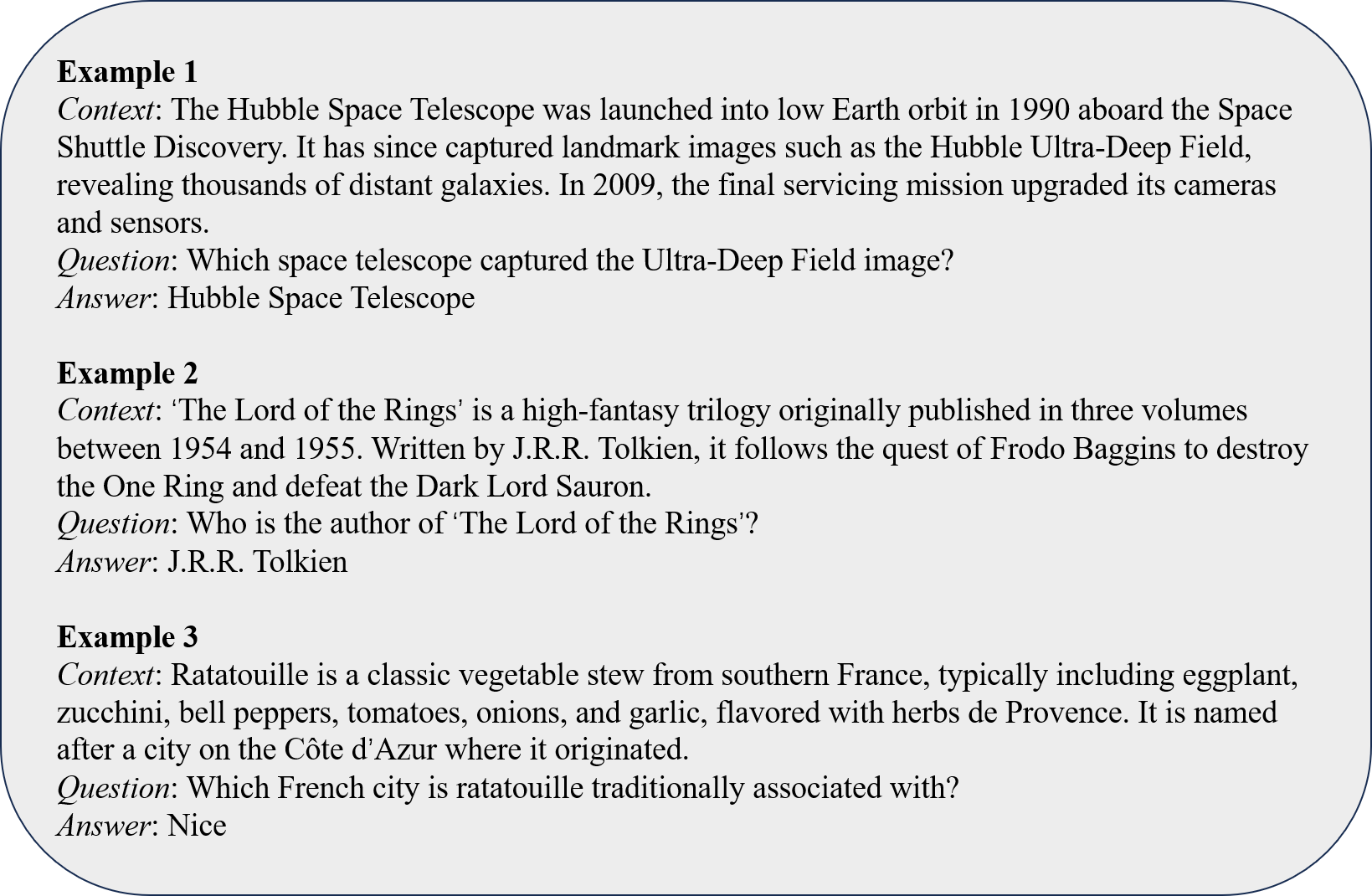}%
  }
  \caption{Prompting examples used in the SQuAD \texttt{dev-v1.1} task under different few-shot settings. Zero-shot uses no demonstrations, one-shot includes only Example~1, and three-shot includes all three examples.}
  \label{fig:shots_examples}
\end{figure}

Figure~\ref{fig:shots_examples} shows the chain-of-thought prompting examples we use for the SQuAD \texttt{dev-v1.1} task. In the \textbf{zero-shot} setting, no demonstrations are provided. The \textbf{one-shot} setting includes only Example~1, while the \textbf{three-shot} setting incorporates all three examples.

\begin{figure}[h]
  \centering
  \resizebox{\linewidth}{!}{%
    \includegraphics{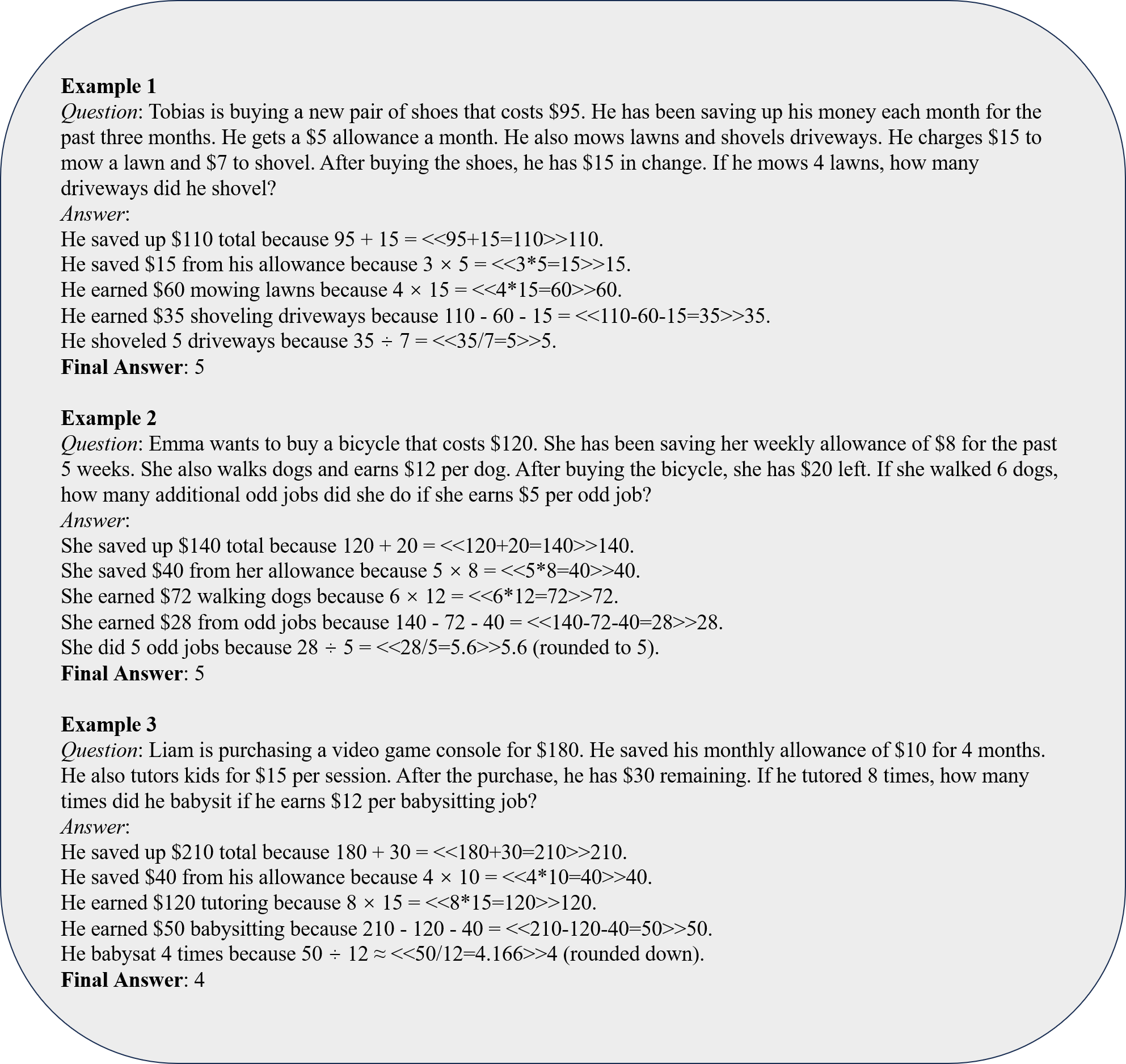}%
  }
  \caption{Prompting examples used in different few-shot settings for the GSM8K task, adapted to arithmetic reasoning.}
  \label{fig:gsm8k_shots}
\end{figure}

Figure~\ref{fig:gsm8k_shots} shows the chain-of-thought demonstrations used for the GSM8K task. Similarly, the \textbf{zero-shot} configuration contains no examples, the \textbf{one-shot} configuration includes only the first example, and the \textbf{three-shot} configuration includes all three. These prompts are used to evaluate the effect of demonstration count on arithmetic reasoning performance.

\section{Analysis on Clustering and Representative Selection}
\label{appendix:5}

As shown in Table~\ref{tab:clustering-representative-squad}, while different clustering algorithms (Greedy, K-Means++, Agglomerative, Spectral) yield nearly identical accuracies, the representative selection strategy makes a substantial difference. Specifically, choosing the first-in-cluster answer consistently outperforms alternatives such as selecting the cluster centroid or the maximum-confidence path. This confirms that index ordering plays a crucial role in GCoT-decoding, and that a greedy clustering scheme combined with first-in-cluster selection is both efficient and effective.

\begin{table}[htbp]
\centering
\caption{Accuracy comparison of clustering methods and representative choices on SQuAD v1.1.}
\label{tab:clustering-representative-squad}
\renewcommand{\arraystretch}{1.05}
\resizebox{\columnwidth}{!}{%
\begin{tabular}{p{2.7cm} p{3.5cm} c c}
\toprule
\textbf{Category} & \textbf{Method} & \textbf{Gemma-7B} & \textbf{Llama-3.1-8B} \\
\midrule
\multirow{4}{*}{Clustering}
& Greedy Clustering       & 54.6 & \textbf{67.2} \\
& K-Means++               & 54.4 & 66.9 \\
& Agglomerative (Ward)    & \textbf{54.7} & 67.1 \\
& Spectral Clustering     & 54.5 & 67.0 \\
\midrule
\multirow{3}{*}{Representative}
& First-in-Cluster        & \textbf{54.6} & \textbf{67.2} \\
& Cluster Centroid        & 47.8 & 60.4 \\
& Max-Conf                & 48.2 & 60.9 \\
\bottomrule
\end{tabular}%
}
\end{table}

\section{Choice of the number of backtracking}
\label{appendix:6}

\begin{figure}[htbp]
  \centering
  \includegraphics[width=0.42\textwidth]{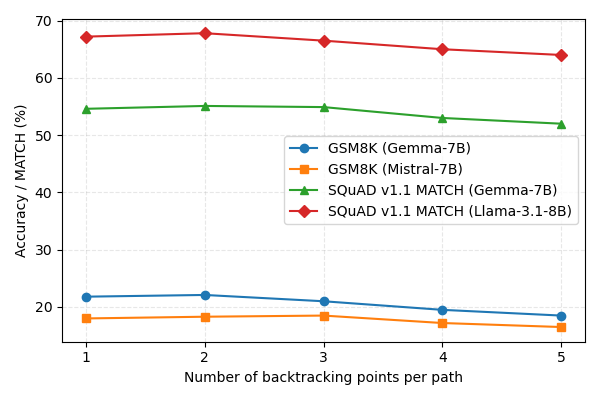}
  \caption{Effect of the maximum number of backtracking points per path under a fixed overall path budget.}
  \label{fig:multi_backtracking}
\end{figure}

We find that CoT errors tend to have early turning points: as soon as the model commits to a wrong semantic decision (Table~\ref{tab:error-entrenchment}), the token-level confidence exhibits a sharp local drop, and subsequent tokens mostly elaborate on this misconception rather than correcting it. In these cases, backtracking at the first confidence valley is typically sufficient to redirect the reasoning towards a different, potentially correct branch. From an efficiency perspective, allowing multiple backtracking points per path under a fixed path budget significantly increases decoding cost and complicates how to trade off early vs.\ late corrections, so we adopt a simple one-shot backtracking rule as a pragmatic accuracy–efficiency compromise.

Figure~\ref{fig:multi_backtracking} summarizes this ablation by varying the maximum number of backtracking points per path from 1 to 5: performance improves slightly from 1-back to 2-back, stays roughly flat around 3-back, and then drops noticeably at 4 and 5. This pattern indicates that limited extra backtracking offers only marginal gains, while aggressive multi-backtracking quickly hurts both accuracy and efficiency, supporting our choice of a single-shot local-minima strategy.

\begin{table*}[htbp]
  \centering
  \resizebox{\textwidth}{!}{%
  \begin{tabular}{lccccccccc}
    \toprule
    \multirow{2}{*}{Method}
    & \multicolumn{3}{c}{GSM8K (Acc.)}
    & \multicolumn{3}{c}{MultiArith (Acc.)}
    & \multicolumn{3}{c}{Sports Understanding (Acc.)} \\
    \cmidrule(lr){2-4} \cmidrule(lr){5-7} \cmidrule(lr){8-10}
    & Mistral-7B & Gemma-7B & Llama-3.1-8B
    & Mistral-7B & Gemma-7B & Llama-3.1-8B
    & Mistral-7B & Gemma-7B & Llama-3.1-8B \\
    \midrule
    GCoT-decoding + SpanAlign (Last)
    & 10.7 & 15.4 & 34.0
    & 16.8 & 19.7 & 69.3
    & 48.0 & 67.2 & 52.0 \\
    GCoT-decoding + SpanAlign (Mean)
    & 10.2 & 14.9 & 33.5
    & 16.1 & 19.0 & 68.5
    & 47.1 & 66.3 & 51.4 \\
    \bottomrule
  \end{tabular}%
  }
  \caption{Comparison between using only the last aligned answer span (SpanAlign~(Last)) and averaging over all aligned spans (SpanAlign~(Mean)).}
  \label{tab:spanalign_last_mean}
\end{table*}

\section{Effect of answer-extraction templates}
\label{appendix:template_ablation}

In Ewe use a short continuation template (e.g., ``So the answer is ...'') purely as an answer-extraction marker after the model has already produced a full chain-of-thought reasoning trace. To verify that GCoT-decoding does not depend on the specific wording of this marker, we evaluate several semantically equivalent templates on SQuAD v1.1 with Gemma-7B, while keeping all other components fixed (Table~\ref{tab:template_ablation}).

\begin{table}[htbp]
  \centering
  \resizebox{0.5\textwidth}{!}{%
  \begin{tabular}{lc}
    \toprule
    Template & SQuAD v1.1 MATCH (Gemma-7B) \\
    \midrule
    ``So the answer is \dots''         & 54.6 \\
    ``Therefore, the answer is \dots'' & 54.5 \\
    ``Final answer:''                  & 54.3 \\
    \bottomrule
  \end{tabular}%
  }
  \caption{Ablation on answer-extraction templates for GCoT-decoding on SQuAD v1.1.}
  \label{tab:template_ablation}
\end{table}

The variation across templates is within 0.3 absolute MATCH points, which is negligible compared to the gains obtained by switching from greedy or vanilla CoT-decoding to GCoT on the same benchmark. This supports our claim that GCoT-decoding does not hinge on a specific wording of the answer-extraction template.

\section{Embedding model ablation for semantic clustering}
\label{appendix:embed_ablation}

GCoT-decoding uses an off-the-shelf sentence embedding model to perform greedy semantic clustering over candidate paths. To assess the sensitivity of this module to the choice of embedding space, we fix the rest of the framework and only vary the embedding model, comparing MiniLM, MPNet-base, and E5-small on SQuAD v1.1 and Auto-Categorization (Table~\ref{tab:embed_ablation}).

\begin{table}[htbp]
  \centering
  \resizebox{0.5\textwidth}{!}{%
  \begin{tabular}{lcccc}
    \toprule
    Setting
    & \makecell[c]{SQuAD v1.1\\BLEU}
    & \makecell[c]{SQuAD v1.1\\MATCH}
    & \makecell[c]{Auto-cat\\BLEU}
    & \makecell[c]{Auto-cat\\MATCH} \\
    \midrule
    GCoT + MiniLM      & 10.0 & 67.2 & 10.6 & 30.5 \\
    GCoT + MPNet-base  &  9.8 & 66.7 & 10.4 & 30.3 \\
    GCoT + E5-small    & 10.1 & 67.0 & 10.5 & 30.4 \\
    \bottomrule
  \end{tabular}%
  }
  \caption{Embedding model ablation for the semantic clustering module in GCoT-decoding.}
  \label{tab:embed_ablation}
\end{table}

Across all settings, the variation in BLEU and MATCH is within 0.5 absolute points, suggesting that the greedy clustering module is relatively insensitive to the specific off-the-shelf embedding model used, as long as it provides a reasonable semantic similarity signal. This matches our design goal of treating semantic clustering as a conservative, pluggable enhancement over simple max-path selection.

\section{SpanAlign ablation: last vs.\ mean alignment}
\label{appendix:spanalign_ablation}

In Section~\ref{sec:confidence}, we use an LCS-based \textsc{SpanAlign} module to compare answer segments across different paths. When the same answer phrase appears multiple times in a reasoning trace, our default implementation scores only the terminal aligned segment (``SpanAlign~(Last)''). To check whether averaging over all aligned segments could be preferable, we compare this default against a variant that averages confidence across all occurrences (``SpanAlign~(Mean)'') on GSM8K, MultiArith, and Sports Understanding (Table~\ref{tab:spanalign_last_mean}).

Across all three datasets and models, using the final occurrence of the aligned answer span is at least as reliable as averaging over all occurrences, and often slightly better.

\section{Qualitative example of early path backtracking}
\label{app:backtracking_case_study}
\begin{table}[ht]
\centering
\caption{An example of path backtracking. The underlined segments indicate the answers targeted by the decoding paths, while the highlighted portions show the content generated after backtracking. “plays”, “defensive”, and “.” are the three local minima in Path1.}
\label{tab:error-entrenchment}
\scriptsize
\begin{tabular}{>{\raggedright\arraybackslash}p{0.95\columnwidth}}
\toprule
\textbf{Question}: \textit{What position does Von Miller play?}

\textbf{Path1(×)}: \textit{Von Miller plays(0.0877) \ul{\textbf{defensive(0.0921) end}} position .(0.1980)}

\textbf{Path2(\checkmark)}: \textit{Von plays the \ul{\textbf{outside linebacker}} \hl{position on his team .}}

\textbf{Path3(×)}: \textit{Von Miller plays the \ul{\textbf{defensive end}} \hl{role for his team and is known for his pass rushing ability .}}\\
\bottomrule
\end{tabular}
\end{table}

We provide a qualitative example in Table~\ref{tab:error-entrenchment} to illustrate early error correction in the decoding process.
In Path1, the incorrect answer “defensive end” emerges after three local minima.
Branching before the first error token (e.g., at “plays”) allows effective correction, as in Path2, which leads to the correct answer “linebacker.”
In contrast, branching after the error fragment has formed, as in Path3, fails to revise the mistake—once embedded, the error resists recovery.
This highlights the importance of early branching before erroneous spans are committed.

\end{document}